\documentclass{article}
\usepackage[bottom]{footmisc}
\usepackage{float}
\usepackage{sidecap}
\usepackage{multirow}
\usepackage{parskip}

\usepackage[margin=1.25in]{geometry}
\usepackage[utf8]{inputenc} % allow utf-8 input
\usepackage[T1]{fontenc}    % use 8-bit T1 fonts
\usepackage{url}            % simple URL typesetting
\usepackage{booktabs}       % professional-quality tables
\usepackage{amsfonts}       % blackboard math symbols
\usepackage{nicefrac}       % compact symbols for 1/2, etc.
\usepackage{microtype}      % microtypography
\usepackage{kpfonts}
\usepackage{graphicx}
\usepackage{multicol}
\usepackage{enumitem}
\usepackage{wrapfig}
\usepackage{natbib}
\usepackage{smile}

\usepackage[small]{caption}
\usepackage[table]{xcolor}
\usepackage{listings}
\usepackage{xcolor}

\usepackage{tcolorbox}
\usepackage{caption}

\usepackage{array}

\definecolor{codeblue}{rgb}{0,0.28,0.67}

\usepackage{booktabs, threeparttable, xcolor, tabularx}
\definecolor{nmHeader}{HTML}{2B2B2B}
\definecolor{nmHeadTxt}{HTML}{FFFFFF}
\definecolor{nmBand}{HTML}{F7F7F7}
\definecolor{nmSubhdr}{HTML}{E9E9E9}
\newcolumntype{Y}{>{\centering\arraybackslash}X}

\definecolor{TableHeader}{HTML}{ECEFF3}
\definecolor{RowTint}{HTML}{F8FAFC}
\renewcommand{\arraystretch}{1.3}
\newcommand{\NA}{\textemdash}

\usepackage{hyperref}
\hypersetup{
    colorlinks,
    linkcolor={codeblue},
    citecolor={codeblue},
    urlcolor={codeblue}
}

\usepackage[scaled]{beramono}

\usepackage{lipsum}
\usepackage{listings}

\lstdefinestyle{mystyle}{
    escapechar=\%,
    keywordstyle=\color{codeblue},
    basicstyle=\ttfamily\small,
    breakatwhitespace=false,         
    breaklines=true,                 
    captionpos=b,                    
    keepspaces=true,                 
    numbers=left,                    
    numbersep=5pt,                  
    showspaces=false,                
    showstringspaces=false,
    showtabs=false,                  
    tabsize=2
}
\usepackage{amsthm}
\usepackage{mathtools}

\usepackage{setspace}

\newcommand\eg {{\it e.g., }}

\newcommand\etc {{etc.}}

\newcommand\ie {{\it i.e., }}
\newcommand\vs {{\it vs. }}

\newcommand\freeresponsestart {{\textless free\_response\textgreater}}
\newcommand\freeresponseend {{\textless /free\_response\textgreater}}
\newcommand\descriptionstart {{\textless description\textgreater}}
\newcommand\descriptionend {{\textless /description\textgreater}}
\newcommand\multiplechoicestart {{\textless multiple\_choice\textgreater}}
\newcommand\multiplechoicend {{\textless /multiple\_choice\textgreater}}
\newcommand\qstart {{\textless q\textgreater}}
\newcommand\qend {{\textless /q\textgreater}}
\newcommand\astart {{\textless a\textgreater}}
\newcommand\aend {{\textless /a\textgreater}}

\definecolor{headerblue}{RGB}{200,230,255}
\definecolor{sectionorange}{RGB}{255,200,150}

\title{\Large
{Mammo-FM: Breast-specific foundational model for Integrated Mammographic Diagnosis, Prognosis, and Reporting}
}
\author{
\normalsize
Shantanu Ghosh$^{\dagger,1}$,
Vedant Parthesh Joshi$^{*, 3}$,
Rayan Syed$^{*, 1}$, 
Param Budhraja$^{*, 1}$,\\
\normalsize
Aya Kassem$^{*, 1}$, 
Katelyn C. Morrison$^{10}$, 
Alex Tang$^{*, 9}$, 
Ho Cheung Aiden Wong$^{2}$, 
Abhishek Varshney$^{2}$, \\
\normalsize
Payel Basak$^{4}$,
Weicheng Dai$^{1}$,
Judy Wawira Gichoya$^{5}$, 
Hari M. Trivedi$^{5}$, 
Imon Banerjee$^6$,\\
\normalsize
Shyam Visweswaran$^{8}$,
Clare B. Poynton$^7$,
Kayhan Batmanghelich$^{1}$\\ 
\\
\normalsize
\normalsize $^{1}$ Department of Electrical and Computer Engineering, Boston University, Boston, MA, USA\\
\normalsize $^{2}$ Department of Computer Science, Boston University, Boston, MA, USA\\
\normalsize $^{3}$ Data Science, Analytics and Engineering, Arizona State University, Tempe, AZ, USA\\
\normalsize $^{4}$ Center for Regenerative Medicine, Boston University Medical Campus, Boston, MA, USA\\
\normalsize $^{5}$ Department of Radiology, Emory University, Atlanta, GA, USA\\
\normalsize $^{6}$ Department of Radiology, Mayo Clinic, Phoenix, AZ, USA\\
\normalsize $^{7}$ Chobanian \& Avedisian School of Medicine,  Boston, MA, USA\\
\normalsize $^{8}$Department of Biomedical Informatics, University of Pittsburgh, Pittsburgh, PA,
USA\\
\normalsize $^{9}$Department of Computer Science, Dartmouth College, Hanover, NH,
USA\\
\normalsize $^{10}$Human-Computer Interaction, Carnegie Mellon University, Pittsburgh, PA,
USA\\
\normalsize  $^{*}$ These authors contributed equally to this work.\\
\normalsize  $^{\dagger}$ Corresponding author(s). E-mail(s): \href{mailto:shawn24@bu.edu}{shawn24@bu.edu}\\
\normalsize  Contributing authors: 
\normalsize  \{
\href{mailto:rsyed@bu.edu}{rsyed},
\href{mailto:paramb@bu.edu}{paramb}
\href{mailto:ahcmwong@bu.edu}{ahcmwong}
\href{mailto:ayak@bu.edu}{ayak},
\href{mailto:avarshn@bu.edu}{avarshn},
\href{mailto:payelb@bu.edu}{payelb},
\href{mailto:wd2119@bu.edu}{wd2119},
\href{mailto:batman@bu.edu}{batman}
\}@bu.edu; \\
\normalsize \href{mailto:vjoshi22@asu.edu}{vjoshi22@asu.edu}; 
\normalsize \href{mailto:alex.z.tang.28@dartmouth.edu}{alex.z.tang.28@dartmouth.edu}; 
\normalsize \href{mailto:kcmorris@andrew.cmu.edu}{kcmorris@andrew.cmu.edu}; \\
\normalsize  \href{mailto:judywawira@emory.edu}{judywawira@emory.edu};
\normalsize  \href{mailto:hari.trivedi@emory.edu}{hari.trivedi@emory.edu};
\normalsize  \href{mailto:banerjee.imon@mayo.edu}{banerjee.imon@mayo.edu};\\
\normalsize  \href{mailto:shv3@pitt.edu}{shv3@pitt.edu}; 
\href{mailto:clare.poynton@bmc.org}{clare.poynton@bmc.org}
}

\date{}
\begin{document}

\maketitle

\vspace{-1.5ex}
\begin{abstract}
\noindent

Breast cancer is one of the leading causes of death among women worldwide. We introduce \textbf{Mammo-FM}, the first foundation model specifically for mammography, pretrained on the largest and most diverse dataset to date—140,677 patients (821,326 mammograms) across four U.S. institutions. Mammo-FM provides a unified foundation for core clinical tasks in breast imaging, including cancer diagnosis, pathology localization, structured report generation, and cancer risk prognosis within a single framework. Its alignment between images and text enables both visual and textual interpretability, improving transparency and clinical auditability, which are essential for real-world adoption. We rigorously evaluate Mammo-FM across diagnosis, prognosis, and report-generation tasks in in- and out-of-distribution datasets. Despite operating on native-resolution mammograms and using only one-third of the parameters of state-of-the-art generalist FMs, Mammo-FM consistently outperforms them across multiple public and private benchmarks. These results highlight the efficiency and value of domain-specific foundation models designed around the full spectrum of tasks within a clinical domain and emphasize the importance of rigorous, domain-aligned evaluation. The code and checkpoint is available at \href{here}{https://github.com/batmanlab/Mammo-FM}

\end{abstract}
\vspace{5ex}

\section{Introduction}

In 2022, breast cancer accounted for 2.3 million new cases and 666,000 deaths globally, making it the most diagnosed cancer worldwide~\cite{bray2024global}. 
Artificial Intelligence (AI) has shown promise in assisting radiologists with breast cancer diagnosis~\cite{eisemann2025nationwide}. 
However, improving life expectancy hinges on early detection, which requires early and routine screening. 
Screening-based early detection reduces breast cancer mortality by 25\% over 10 years~\cite{duffy2020effect, trentham2024collaborative}. 
Yet large-scale screening substantially increases radiologist workload and may exacerbate diagnostic errors~\cite{glover2017quantifying, stec2018systematic, kwee2021workload}. 
AI systems can help mitigate these challenges by automating radiologist report generation to reduce non-interpretive burden, and by improving risk stratification to prioritize high-risk patients~\cite{donnelly2024asymmirai}. 
However, developing separate AI models for each task is inefficient, requiring distinct datasets and often reducing generalizability due to smaller sample size. 
Foundation models (FM) offer a unified solution: large-scale pretraining can produce robust, generalizable mammographic representations that support diverse downstream clinical tasks. 
Building on this principle, this paper introduces a mammography-specific FM pretrained on the largest multi-institutional dataset to date, integrating diagnosis, risk prediction, and report generation.

Multimodal FMs pretrained on large-scale image–text datasets have achieved remarkable success in computer vision~\cite{radford2021learning, zhang2020contrastive, huang2021gloria, wu2023medklip, wang2022medclip} and, more recently, in medical imaging~\cite{wang2025large}. 
Collecting paired images and reports at scale is often more feasible than obtaining fine-grained expert annotations. 
The premise of multimodal FMs, and vision–language models in particular, is that large-scale pretraining enables models to learn semantically rich visual representations that can be adapted efficiently to downstream tasks. 
This paradigm aligns naturally with screening mammography. Radiology reports contain key pathological indicators, \eg masses and microcalcifications, which are strong risk factors for breast cancer~\cite{boyd2007mammographic, kim2022microcalcifications, azam2021mammographic}. 
These clinical texts can bootstrap a model’s ability of both diagnostic and prognostic features. 
FMs also alleviate the scarcity of high-quality annotated data in medical imaging, where detailed labels are costly to obtain. 
Large-scale pretraining enables strong zero-shot and linear-probe performance, supporting efficient transfer to new tasks and clinical sites with minimal annotation. 
Pretraining on heterogeneous datasets further improves out-of-distribution generalization across institutions. 
Finally, image–text alignment provides a direct path to interpretability by grounding visual evidence in natural language, supporting clinical auditing and trust~\cite{varma2023villa, zhong2022regionclip}.

One potential strategy is to construct mammography vision–language models using so-called \emph{generalist FMs} (\eg MedGemma~\cite{sellergren2025medgemma}) that are trained on large-scale medical image data but not specifically on mammograms, or to fine-tune FMs pretrained on other imaging modalities, \eg chest X-rays~\cite{huang2021gloria, wu2023medklip, wang2022medclip}. However, these approaches have fundamental limitations. 
Generalist models often do not operate at native mammography resolution, missing subtle yet critical findings such as microcalcifications. 
Moreover, models trained on heterogeneous clinical data or on modalities unrelated to mammography lack the domain specificity required for high-fidelity breast imaging tasks. 
Our experiments show that these models exhibit poor data efficiency and weak zero-shot performance in the high-resolution, fine-grained mammography setting. 
These limitations motivate the need for a domain-specific FM guided by four core desiderata: 
1) operating directly on native high-resolution mammograms; 
2) handling the realities of clinical data, \ie heterogeneous data with missing reports, partial labels, or incomplete views; 
3) enabling the main tasks in breast cancer screening, including diagnosis, prognosis, pathology localization, and radiology report generation; and 
4) maintaining robustness to dataset shift by generalizing reliably to unseen institutions and patient demographics.

\begin{figure}[p]
    \centering
    \includegraphics[width=0.92\textwidth]{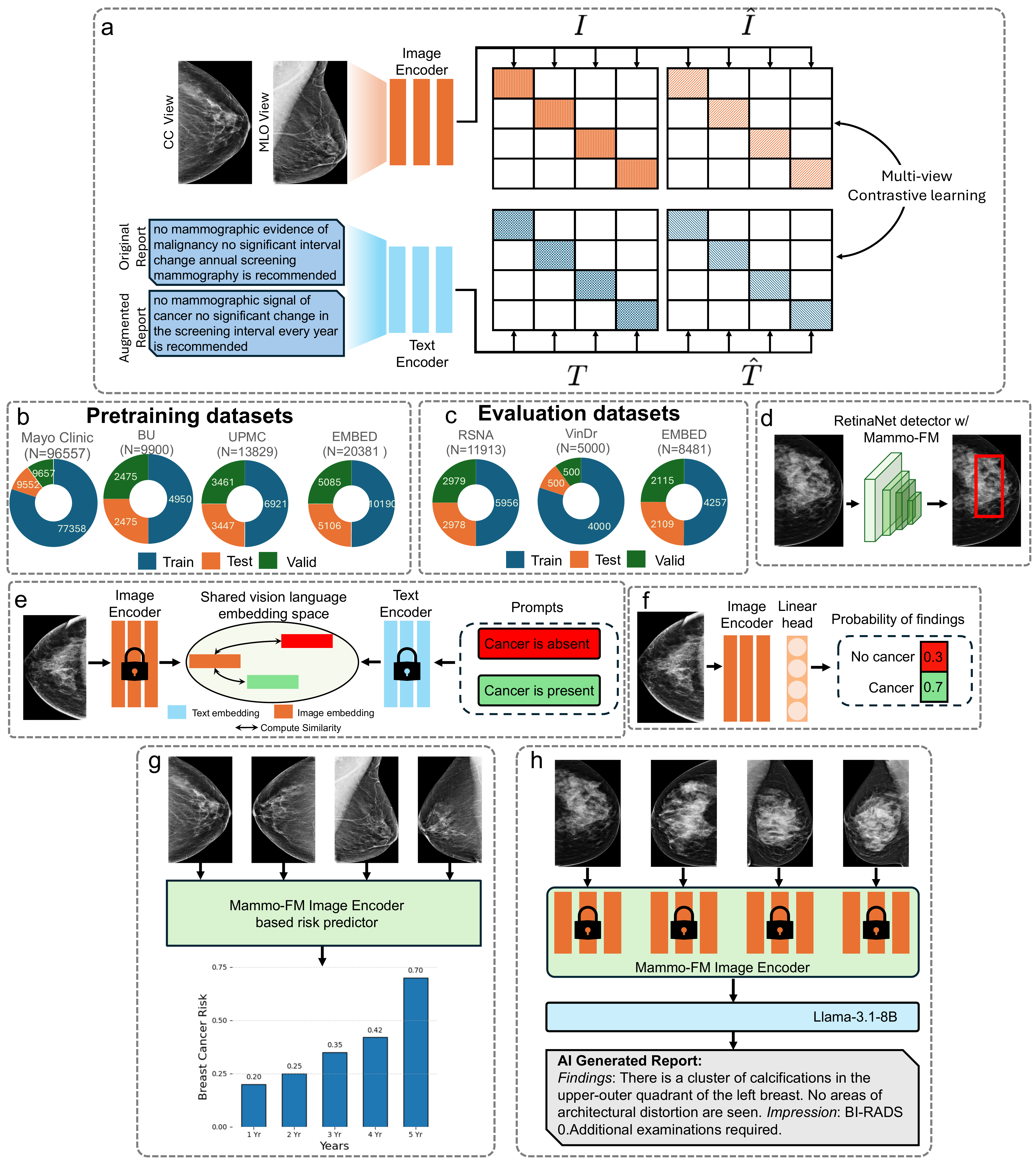}
    \vspace{-0.8em}
    \caption{\textbf{Overview of the Mammo-FM foundation model and downstream applications.} \textbf{a.} Schematic of the Mammo-FM framework. High-resolution mammographic views (CC, MLO) and paired radiology reports are jointly used for multi-view contrastive pretraining, aligning image and text representations within a shared embedding space (see Methods). \textbf{b.} Composition of multi-institutional pretraining datasets from the Mayo Clinic, BU, UPMC, and EMBED cohorts. \textbf{c.} Composition of evaluation datasets used to benchmark diagnostic and prognostic generalization across in-distribution (ID; EMBED) and out-of-distribution (OOD; RSNA, VinDr) settings. \textbf{d.} Pathology localization by integrating the Mammo-FM vision encoder into a RetinaNet detector. \textbf{e.} Zero-shot classification with Mammo-FM. Image and text encoders map mammograms and prompts into a shared space. Similarity between embeddings enables classification without fine-tuning. \textbf{f.} Findings classification under either linear-probe or full fine-tuning settings, where a linear classifier is attached to Mammo-FM image encoder, quantifying the quality of the learned visual representations. \textbf{g.} Interpretable risk prediction built using frozen Mammo-FM encoders across CC and MLO views for 1–5 year risk estimation. \textbf{h.} Automated report generation (Mammo-GRG) combining Mammo-FM with a Llama-3.1-8B language model to produce clinically grounded radiology reports.}  
    \label{fig:schematic}
\end{figure}

\begin{table*}[t]
\centering
\footnotesize
\begin{threeparttable}
\caption{Summary characteristics of the EMBED, UPMC, and Mayo Clinic datasets used to pre-train Mammo-FM. Note that the Mammo-GRG pretraining also leverages the same UPMC dataset. All entries represent patient-level characteristics. }
\label{tab:main_dataset_char}
\renewcommand{\arraystretch}{1.15}
\setlength{\tabcolsep}{5pt}

\begin{tabularx}{\textwidth}{l l Y Y Y}
\rowcolor{nmHeader}
\textbf{\color{nmHeadTxt} Characteristic} &
\textbf{\color{nmHeadTxt} Category} &
\textbf{\color{nmHeadTxt} EMBED (N=20,381)} &
\textbf{\color{nmHeadTxt} UPMC (N=13{,}829)} &
\textbf{\color{nmHeadTxt} Mayo Clinic (N=96{,}567)} \\

\midrule

\multirowcell{3}[-.2ex][l]{\cellcolor{nmSubhdr}\textbf{BI-RADS}} &
0 & 5{,}476 (15.4\%) & 2{,}125 (15.4\%) & \NA \\
& 1 & 10{,}822 (31.7\%) & 4{,}376 (31.7\%) & \NA \\
& 2 & 4{,}083 (52.9\%) & 7{,}328 (52.9\%) & \NA \\
\addlinespace[3pt]\midrule

\multirowcell{3}[-.2ex][l]{\cellcolor{nmSubhdr}\textbf{Age Group}} &
$<$ 50 years & 956 (9.6\%) & 1{,}326 (9.6\%) & 2{,}476 (2.4\%) \\
& 50–60 years & 11{,}163 (44.6\%) & 6{,}163 (44.6\%) & 45{,}513 (47.3\%) \\
& $>$ 60 years & 8{,}262 (45.8\%) & 6{,}340 (45.8\%) & 48{,}578 (50.4\%) \\
\addlinespace[3pt]\midrule

\multirowcell{4}[-.2ex][l]{\cellcolor{nmSubhdr}\textbf{Breast Density}} &
Almost entirely fatty (A) & 3{,}113 (22.5\%) & 6{,}159 (22.5\%) & 9{,}712 (10.0\%) \\
& Scattered fibroglandular (B) & 5{,}712 (41.3\%) & 5{,}233 (41.3\%) & 47{,}370 (49.0\%) \\
& Heterogeneously dense (C) & 3{,}801 (27.5\%) & 6{,}765 (27.5\%) & 34{,}861 (36.1\%) \\
& Extremely dense (D) & 1{,}203 (8.7\%) & 2{,}224 (8.7\%) & 4{,}624 (4.9\%) \\
\addlinespace[3pt]\midrule

\multirowcell{4}[-.2ex][l]{\cellcolor{nmSubhdr}\textbf{Race}} &
Black & 8{,}580 (42.1\%) & \NA & 893 (0.9\%) \\
& White & 8{,}193 (40.2\%) & \NA & 92{,}287 (95.3\%) \\
& Asian & 1{,}305 (6.4\%) & \NA & 1{,}713 (1.7\%) \\
& Other & 2{,}303 (11.3\%) & \NA & 1{,}674 (1.7\%) \\
\addlinespace[3pt]\midrule

\multirowcell{5}[-.2ex][l]{\cellcolor{nmSubhdr}\textbf{Manufacturer}} &
HOLOGIC, Inc. & \NA & 13{,}443 (96.7\%) & 90{,}351 (93.5\%) \\
& LORAD & \NA & 287 (2.1\%) & 1{,}359 (1.5\%) \\
& SIEMENS & \NA & 99 (0.7\%) & 5 ($<$0.1\%) \\
& Visage & \NA & \NA & 20 ($<$0.1\%) \\
& Unknown & \NA & \NA & 4{,}832 (5.0\%) \\

\bottomrule
\end{tabularx}

\begin{tablenotes}[flushleft]\footnotesize
\item \textbf{Note:} Missing entries are indicated by \NA. BI-RADS categories follow standard clinical grading; percentages are computed relative to total patient count.
\end{tablenotes}
\end{threeparttable}
\end{table*}

\begin{table*}[t]
\centering
\footnotesize
\begin{threeparttable}
\caption{Characteristics of BU datasets. It consists of 54,776 screening images from 9,900 patients and 13,480 exams with complete data, that includes an associated radiology report and followup data on clinical outcomes. It is one of the pretraining datasets used to pre-train Mammo-FM and Mammo-GRG. Among the 1,041 total cancer cases, the number of cases occurring within years 1–5 are 283, 212, 205, 188, and 153, respectively. Note that, all entries represent exam-level characteristics.}
\label{tab:bu_dataset_characteristics}
\renewcommand{\arraystretch}{1.15}
\setlength{\tabcolsep}{5pt}

\begin{tabularx}{\textwidth}{l l Y Y}
\rowcolor{nmHeader}
\textbf{\color{nmHeadTxt} Characteristic} &
\textbf{\color{nmHeadTxt} Category} &
\textbf{\color{nmHeadTxt} All} &
\textbf{\color{nmHeadTxt} Cases} \\
\midrule

\rowcolor{nmSubhdr}\textbf{Exams} &  & 13{,}480 & 1{,}041 \\
\addlinespace[3pt]\midrule

\multirowcell{4}[-.2ex][l]{\cellcolor{nmSubhdr}\textbf{BI-RADS}} &
 0 & 1{,}333 (9.9\%) & 319 (30.6\%) \\
& 1 & 11{,}062 (82.1\%) & 620 (59.6\%) \\
& 2 & 1{,}053 (7.8\%) & 81 (7.8\%) \\
& Other & 32 (0.2\%) & 21 (2.0\%) \\
\addlinespace[3pt]\midrule

\multirowcell{2}[-.2ex][l]{\cellcolor{nmSubhdr}\textbf{Age at screening}} &
Median (y) & 56 & 62 \\
& Range (y) & 24--87 & 34--87 \\
\addlinespace[3pt]\midrule

\multirowcell{3}[-.2ex][l]{\cellcolor{nmSubhdr}\textbf{Age Group}} &
$<$ 50 y & 3{,}878 (28.7\%) & 114 (10.9\%) \\
& 50--70 y & 7{,}990 (59.3\%) & 747 (71.8\%) \\
& $>$ 70 y & 1{,}617 (12.0\%) & 180 (17.3\%) \\
\addlinespace[3pt]\midrule

\multirowcell{5}[-.2ex][l]{\cellcolor{nmSubhdr}\textbf{Breast Density}} &
 Almost entirely fatty (A) & 3{,}428 (25.4\%) & 190 (18.3\%) \\
& Scattered fibroglandular (B) & 4{,}792 (35.6\%) & 524 (50.3\%) \\
& Heterogeneously dense (C) & 3{,}824 (28.3\%) & 268 (25.7\%) \\
& Extremely dense (D) & 1{,}323 (9.8\%) & 50 (4.8\%) \\
& Unknown & 113 (0.8\%) & 9 (0.9\%) \\
\addlinespace[3pt]\midrule

\multirowcell{6}[-.2ex][l]{\cellcolor{nmSubhdr}\textbf{Race}} &
 Black & 7{,}224 (53.6\%) & 557 (53.5\%) \\
& White & 2{,}573 (19.1\%) & 185 (17.8\%) \\
& Hispanic & 3{,}219 (23.9\%) & 251 (24.1\%) \\
& Asian & 327 (2.4\%) & 31 (3.0\%) \\
& Native American & 31 (0.2\%) & 3 (0.3\%) \\
& Unknown & 106 (0.8\%) & 14 (1.3\%) \\
\midrule

\multirowcell{2}[-.2ex][l]{\cellcolor{nmSubhdr}\textbf{Family history of breast cancer.}} &
 No & 12{,}950 (96.1\%) & 930 (89.3\%) \\
& Yes & 530 (3.9\%) & 111 (10.7\%) \\
\addlinespace[3pt]\midrule

\multirowcell{2}[-.2ex][l]{\cellcolor{nmSubhdr}\textbf{History of breast biopsy}} &
 No prior biopsy & 12{,}901 (95.7\%) & 922 (88.6\%) \\
& Prior biopsy, benign dx unknown & 579 (4.3\%) & 119 (11.4\%) \\
\addlinespace[3pt]

\bottomrule
\end{tabularx}

\begin{tablenotes}[flushleft]\footnotesize
\item \textbf{Note:} Percentages are relative to the row’s population (\textbf{All} or \textbf{Cases}). Cases denotes \textbf{cancer cases}. Abbreviations: dx, diagnosis. Family history of breast cancer denotes cancer in first-degree relative(s).
\end{tablenotes}
\end{threeparttable}
\end{table*}

To address these challenges, we propose \textbf{Mammo-FM}, a vision--language model specifically designed for mammography. Mammo-FM uses dual encoders—an EfficientNet-B5 image encoder and a ModernBERT text encoder finetuned on $\sim$200k radiology reports—to jointly process mammograms and paired reports. A multiview contrastive loss aligns both modalities within a shared embedding space (Methods and Fig.~\ref{fig:schematic}a). Unlike generalist FMs, Mammo-FM operates directly on high-resolution 2D screening mammograms, preserving subtle but clinically essential features. 
To learn robust representations, we pretrain Mammo-FM on a multi-center corpus of 140,677 patients (821,326 screening mammograms) from four U.S. hospitals: UPMC (13,829 patients; 46,433 mammograms), Boston University (9,900 patients; 54{,}776 mammograms), EMBED (20,381 patients; 255,039 mammograms), and Mayo Clinic (96,567 patients; 465,078 mammograms) (see Tab.~\ref{tab:main_dataset_char} and Tab.~\ref{tab:bu_dataset_characteristics}). This represents the largest and most diverse mammography pretraining dataset to date.

Despite using a 30M-parameter vision encoder—far smaller than the 400M-parameter MedSigLIP encoder in MedGemma—Mammo-FM outperforms MedSigLIP in zero-shot and data-efficient settings across public in-distribution (ID) and out-of-distribution (OOD) benchmarks, all at a fraction of the pretraining cost.
Leveraging strong image–text alignment, Mammo-FM supports two challenging clinical applications: the first text-interpretable mammography risk predictor and the first high-resolution, multi-view mammography report generator. 
Mammo-FM transforms state-of-the-art risk prediction models such as MIRAI~\cite{yala2021toward} and AsymMIRAI~\cite{donnelly2024asymmirai} into textually interpretable systems by providing aligned visual and textual features. Motivated by its superior diagnostic performance, we integrate Mammo-FM’s report-aligned vision encoder into both pipelines. This offers clinicians richer decision-support beyond a single risk score. Mammo-FM’s alignment allows us to highlight sentences in the paired radiology report that correspond to the visual cues driving the risk prediction. Mammo-FM–based risk predictors achieve strong performance across ID and public OOD datasets.

We next introduce \textbf{ Mammo Grounded Report Generator (Mammo-GRG)}, a high-resolution, multi-view mammography report generator. Mammo-GRG uses four parallel Mammo-FM image encoders to process all mammography views at full resolution. A two-stage training pipeline—(1) aligning frozen encoders with an LLM (Llama 3.1 8B) using 19,811 study–report pairs, and (2) instruction tuning with 359,746 mammography QA pairs—produces a preliminary report. A final grounding stage fuses this text with zero-shot Mammo-FM findings to ensure factual consistency. 
Standard NLP metrics (\eg BLEU) fail to capture clinically critical errors such as incorrect laterality or BI-RADS scores. We therefore adapt the GREEN~\cite{ostmeier2024green} metric for structured, mammography-specific evaluation using GPT-4o-mini as the LLM judge. Mammo-GRG outperforms generalist models in report fidelity, finding-level accuracy, and OOD robustness. 
Together, Mammo-FM and its downstream applications move mammography AI toward high-resolution, interpretable, and clinically reliable workflows.
\label{sec:intro}

\section{Method}
\subsection{Preprocessing}
We process all mammographic examinations from raw DICOM files using \textit{dicomsdl} package (v0.109.4) of Python library. We standardize pixel values to 8-bit intensity and correct MONOCHROME1 photometric formats. We set values below 40 to zero and remove rows and columns with constant intensity, as they represent the background. These operations isolate the breast region and yield images with an average aspect ratio between 1:1.6 and 1:2. We harmonize orientation using DICOM metadata for laterality, view position, and patient orientation, and we apply horizontal or vertical flips to enforce consistent alignment. We resize the cropped images to $1520 \times 912$ pixels with bilinear interpolation and export them as PNG files while preserving patient and examination identifiers in the directory structure. This pipeline harmonizes acquisitions across devices, eliminates irrelevant background, and preserves diagnostic fidelity essential for large-scale foundation model training.

\subsection{Mammo-FM}
Mammo-FM's pre-training strategy aims to align image and text representations into a shared embedding space. Let $f^I(\cdot)$ and $f^T(\cdot)$ denote the vision and text encoders. Let $(\boldsymbol{x}_i^I, \boldsymbol{x}_i^T)$ be the original image-text pair for a patient $i$. For a batch of size $B$, the model processes these inputs to produce sets of normalized representations, $\mathcal{Z}^I = \{\boldsymbol{z^I}_i\}_{i=1}^B$ and $\mathcal{Z}^T = \{\boldsymbol{z^T}_i\}_{i=1}^B$. The model obtains these representations by projecting the raw encoder outputs through a linear layer to a shared dimension, followed by $\ell_2$ normalization.

Mammo-FM learns this cross-modal alignment through a contrastive objective.  For any two sets of representations, this objective pulls corresponding (`positive') pairs closer in the embedding space while pushing non-corresponding (`negative') pairs apart. This is formulated using a symmetric InfoNCE~\cite{chen2020simple} loss:
\begin{equation}
\label{equ: loss_INFO_NCE}
\mathcal{L}(\mathcal{Z}, \Tilde{\mathcal{Z}}) = - \sum_{\boldsymbol{z} \in \mathcal{Z}} \log \frac{\exp(\langle\boldsymbol{z}, \Tilde{\boldsymbol{z}} \rangle/\tau)}{\sum_{\Tilde{\boldsymbol{z}}' \in \Tilde{\mathcal{Z}}} \exp(\langle\boldsymbol{z}, \Tilde{\boldsymbol{z}}'\rangle/\tau)}
,
\end{equation}
\noindent where $\langle\cdot,\cdot\rangle$ is the cosine similarity function and $\tau$ is a learnable temperature parameter that scales the logits.

To learn more robust and generalizable representations, Mammo-FM utilizes augmented views of both the image and text. Let $(\Tilde{\boldsymbol{x}}_i^I, \Tilde{\boldsymbol{x}}_i^T)$ denote the augmented pair for patient $i$. Processing these augmentations yields two additional sets of representations for each batch: augmented images ($\Tilde{\mathcal{Z}}^I$) and augmented texts ($\Tilde{\mathcal{Z}}^T$). Mammo-FM then extends the contrastive objective using a Multi-View Supervision (MVS) strategy~\cite{li2021supervision}. The MVS objective systematically aligns representations among all four views (original and augmented images and texts) by applying the contrastive loss across all distinct pairs of these representation sets. The total MVS loss is formulated as:
\begin{equation}
\label{equ:loss_MVS_sum}
\mathcal{L}_{\text{MVS}} = \sum_{\mathcal{Z}, \tilde{\mathcal{Z}}\in \{\mathcal{Z}^I, \tilde{\mathcal{Z}}^I, \mathcal{Z}^T, \tilde{\mathcal{Z}}^T\}, \mathcal{Z} \neq \tilde{\mathcal{Z}}} \mathcal{L}(\mathcal{Z}, \tilde{\mathcal{Z}}).
\end{equation}
In our implementation, we down-weight the intra-modal text loss term, $\mathcal{L}(\mathcal{Z}^T, \Tilde{\mathcal{Z}}^T)$, by a factor of 0.5.

\subsubsection*{Instance and Dataset Augmentation}
\paragraph{Instance Augmentation.}  We employ a conditional strategy to create augmented pairs based on the availability of views and report sections for each patient $i$. For images, if both craniocaudal (CC) and mediolateral oblique (MLO) views are present, we treat one as the original image $\boldsymbol{x}^I_i$ and the other as its natural augmentation $\tilde{\boldsymbol{x}}^I_i$. If only one view is available, we designate it as the original and generate its augmentation $\tilde{\boldsymbol{x}}^I_i$ by applying a sequence of standard image transformations, including horizontal and vertical flips, affine transformations (rotation, translation, scaling, and shear), and elastic transforms.

We follow a parallel strategy for text. If a report contains both \texttt{Findings} and \texttt{Impression} sections, we treat one as the original text $\boldsymbol{x}^T_i$ and the other as its augmentation $\tilde{\boldsymbol{x}}^T_i$. If only one section is present, we use it as the original and generate its augmentation using back-translation, following the technique proposed in~\cite{you2023cxr}.

\begin{figure*}[h]
\begin{center}
\centerline{\includegraphics[height=7.5cm]{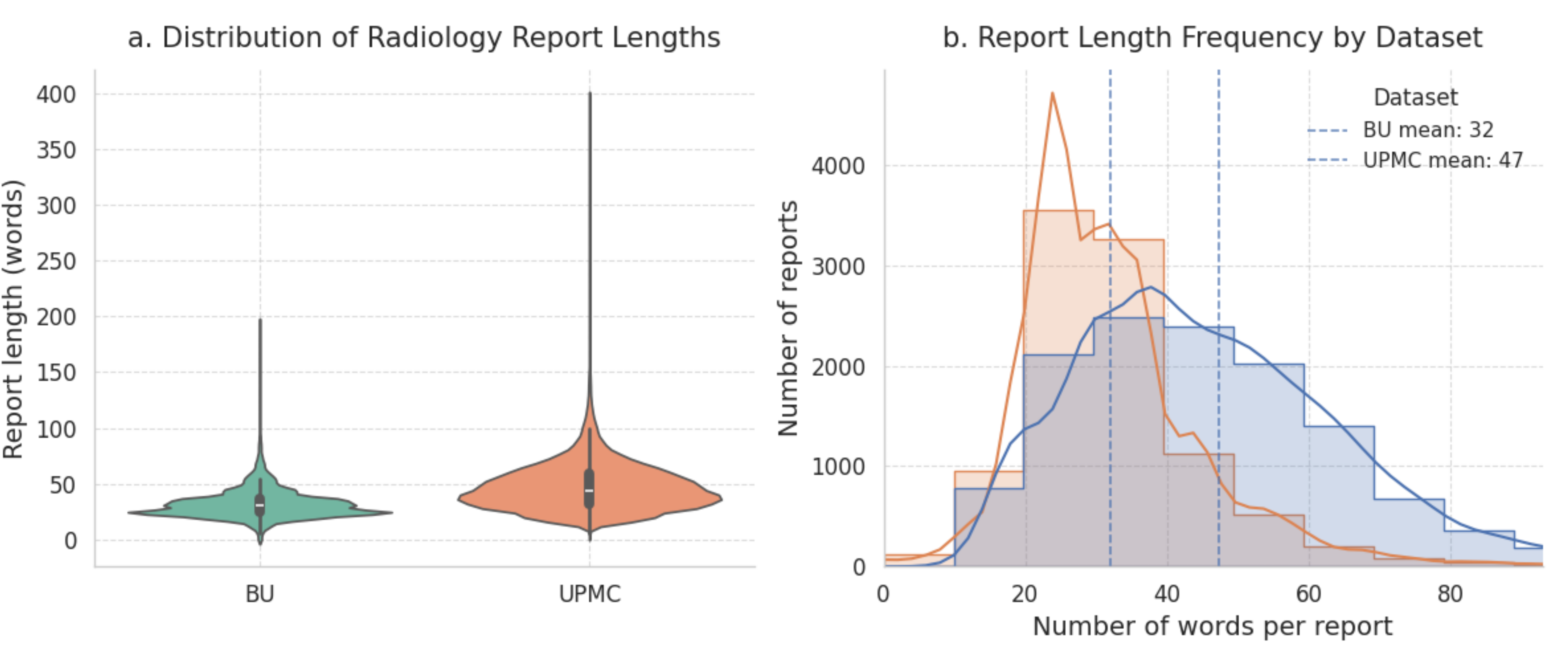}}
\caption{\textbf{Distributions of radiology report lengths across datasets.} \textbf{a.} Violin plots showing the distribution of report word counts for breast imaging datasets from BU and UPMC. Each violin displays the full distribution of report lengths, with internal boxes denoting the interquartile range and median. Reports from UPMC were generally longer and exhibited higher variance compared with BU.
\textbf{b.} Histogram and kernel density estimates (KDEs) comparing report length frequency between BU and UPMC. Dashed vertical lines indicate mean word counts for each dataset (BU mean = 32 words; UPMC mean = 47 words). The distributions highlight systematic differences in reporting style and verbosity across clinical sites.}
\label{fig:app_report_stats}
\end{center}
\end{figure*}

\begin{figure*}[h]
\begin{center}
\centerline{\includegraphics[height=7.5cm]{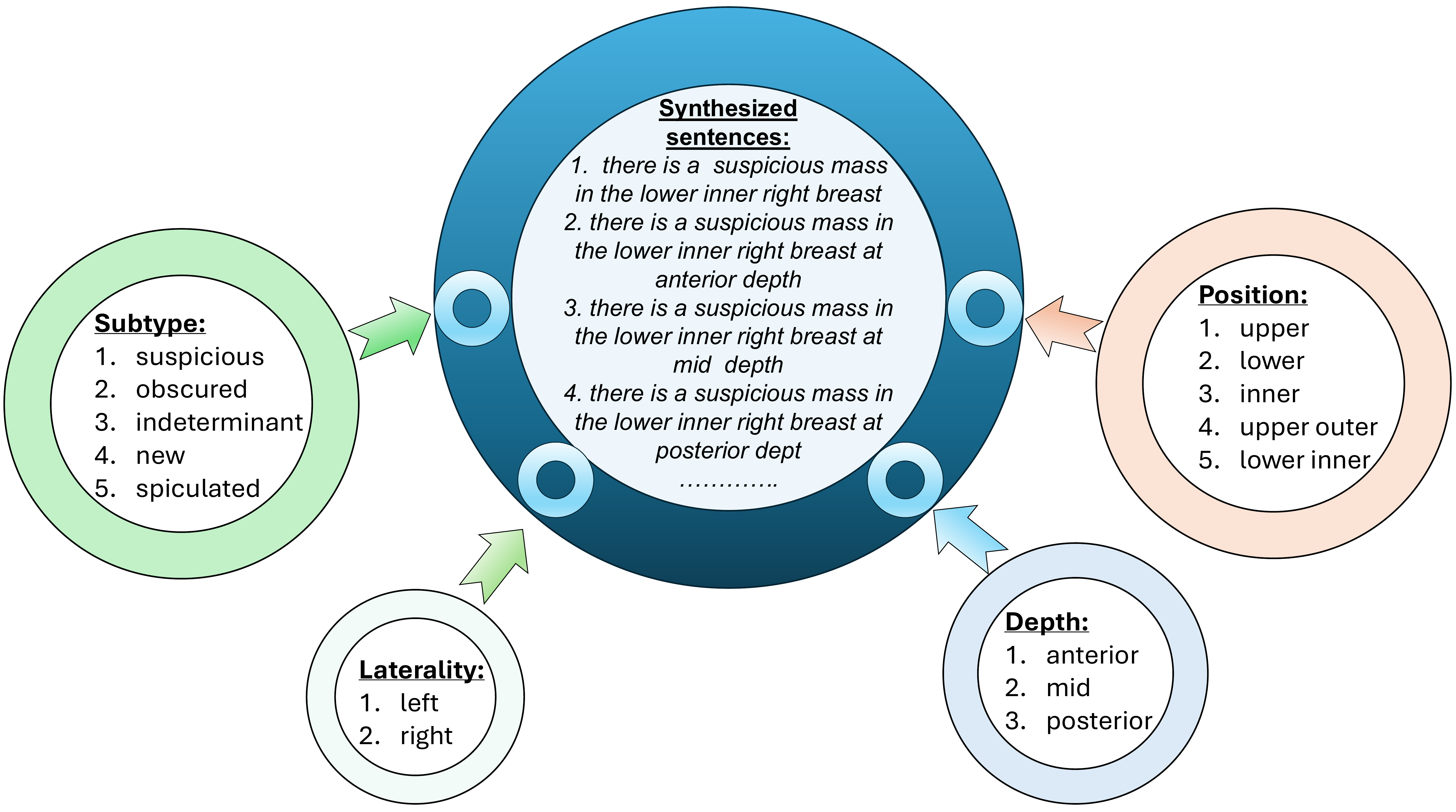}}
\caption{Example of report-like sentence generation for the attribute \textit{mass} labeled positively in the EMBED dataset using the subtypes of mass and position, laterality, and depth of the breast. We include all such prompts in our codebase in detail. We also utilize these prompts to evaluate the zero-shot classification.}
\label{fig:app_image_label_augment}
\end{center}
\end{figure*}

\paragraph{Dataset Augmentation.} To augment the data, we leverage an additional dataset that contains mammograms paired only with structured attributes (\eg \textit{mass, calcification}). We synthesize radiology reports for this dataset to serve as an additional pre-training resource. A board-certified radiologist constructs a set of template sentences that incorporate placeholders for attribute values (e.g., \textit{positive, negative}), subtypes (e.g., \textit{suspicious, obscured}), laterality (\textit{right, left}), and breast location (\textit{upper, lower, anterior}, etc.). For each examination, we synthesize a unique report by randomly selecting a template and populating it with the patient's specific attribute values. We illustrate the process in details in Fig.~\ref{fig:app_image_label_augment}.

\subsubsection*{Architecture}
Mammo-FM employs a dual-encoder architecture to learn a joint embedding space for mammographic images and radiological reports.
For the vision encoder, we use an EfficientNet-B5 model~\cite{tan2019efficientnet}, initialized with weights pre-trained on ImageNet. This architecture provides a strong foundation for learning fine-grained visual features from high-resolution mammograms.
Mammo-FM employs Modern-BERT~\cite{warner2024smarter} as its text encoder, leveraging its domain-specific pretraining on large-scale biomedical corpora to improve representation of medical terminology. Following MedSyn~\cite{xu2024medsyn}, we further fine-tune it on a corpus of 219,451 chest CT radiology reports from the University of Pittsburgh Medical Center (UPMC). The fine-tuning process involves two self-supervised tasks. The first is a Masked Language Modeling (MLM) task, where we randomly mask 15\% of the input tokens and train the model to predict them, enhancing domain-specific lexical knowledge. The second is a Paired Section Prediction task designed to learn high-level semantic consistency. For this task, we swap the \texttt{Impression} section in 50\% of the reports and train the model to predict whether the \texttt{Findings} and \texttt{Impression} sections originate from the same report. For model input, we concatenate the diagnostically critical \texttt{Findings }and \texttt{Impression} sections of each report. After tokenization, the text passes through the fine-tuned Modern-BERT encoder, and we use the contextual embedding of the [CLS] token as the final global text representation. Mammo-FM adopts this fine-tuned text encoder during contrastive pre-training to align with the vision encoder, thereby constructing the shared vision–language embedding space.

\subsubsection*{Training} 
Mammo-FM pretrains over six epochs using a decentralized, cross-institutional protocol designed to preserve data locality while leveraging multi-site diversity. The training begins at a central server hosting the UPMC, BU, and EMBED datasets. After completing one epoch, the model weights are shared with the Mayo Clinic site, which initializes its local model and conducts the second epoch using exclusively Mayo data. The updated weights are then returned to the central server for the third epoch. This alternating schedule continues until all six epochs are completed, enabling full model exposure to heterogeneous data while avoiding raw data transfer. The pretraining pipeline uses mixed-precision computation and optimizes the contrastive objective with AdamW~\cite{loshchilov2017decoupled}, initialized with a learning rate of 5e-5 and weight decay of 1e-4. A cosine-annealing scheduler~\cite{loshchilov2016sgdr} with a linear warm-up during the first epoch modulates the learning rate.  The image augmentation pipeline applies a set of geometric and elastic transformations, including random affine operations—rotation ($\pm20^\circ$), translation (up to 10\%), scaling (0.8--1.2$\times$), and shear ($\pm20^\circ$ -- followed by elastic deformation with $\alpha = 10$ and $\sigma = 15$. The transformation sequence applies with a probability of 1.0 during pretraining.

For text augmentation, Mammo-FM employs a back-translation strategy to ensure semantic consistency while inducing lexical diversity. Specifically, the model translates each report to a pivot language (Italian) from English and back to English using a MarianMT model.

\subsection{Diagnosing Mammographic Findings with the Representation from Mammo-FM}
\textbf{Evaluation setup.}
To rigorously evaluate the quality and generalizability of the representations learned during the pretraining of the multi-institutional Mammo-FM model, we design a comprehensive evaluation framework spanning multiple levels of supervision. Specifically, we evaluate performance in three complementary classification settings: zero-shot (ZS), linear probing (LP), and full fine-tuning (FT). In the ZS setup, we extract image embeddings from the Mammo-FM vision encoder and generate text embeddings from the aligned text encoder. To construct the text embeddings, we convert each target finding \texttt {label} \eg{mass}, into a set of prompts (Fig.~\ref{fig:app_image_label_augment}), generated by a board-certified radiologist. These prompts systematically enumerate: 1) finding subtypes (\eg suspicious, obscured); 2) breast laterality (left, right), 3) depth descriptors (anterior, mid, posterior), and 4) positional qualifiers (upper, lower, central). We encode each prompt independently and average their embeddings to obtain a single representative text embedding per finding category. Finally, we perform the ZS prediction by computing the cosine similarity between the image embedding and this aggregated text embedding, enabling classification without any task-specific training.

For LP and FT, we attach a linear classifier to the Mammo-FM vision encoder. In the LP configuration, the encoder remains frozen and only the linear layer is trained, thereby probing the intrinsic discriminative quality of the learned visual representations for mammographic findings that are explicitly documented in pretraining reports (\eg mass, calcification, architectural distortion, density). In the FT setting, both the vision encoder and the classifier are trained jointly, allowing the model to adapt its representations for the more complex cancer classification task, since malignancy is not explicitly mentioned in the screening mammogram reports used for pre-training.

Each mammographic finding (\eg mass, calcification, density, \etc) is treated as an independent binary classification task for both LP and FT experiments. To further assess data efficiency, both LP and FT models are trained on progressively larger fractions of the training data (10\%, 25\%, 50\%, and 80\%).

Finally, to evaluate the localization capabilities of the learned image features, we integrate the Mammo-FM vision encoder as the backbone of a RetinaNet~\cite{lin2017focal} object detector, assessing its ability not only to classify but also to precisely localize mammographic findings.

\begin{table*}[p]
\centering
\footnotesize
\begin{threeparttable}
\caption{Characteristics of evaluation datasets used to assess Mammo-FM’s image representations. Each dataset is \textbf{publicly} available. Each dataset corresponds to a distinct patient cohort: EMBED (N = 8,481), VinDr (N = 5,000), and RSNA (N = 11,913). Values are reported as patient counts (N) and percentages (\%). These datasets serve as diverse evaluation benchmarks for testing the generalizability and robustness of Mammo-FM’s image–text representations.}
\label{tab:dataset_eval_characteristics}
\renewcommand{\arraystretch}{1.15}
\setlength{\tabcolsep}{5pt}

\begin{tabularx}{\textwidth}{l l Y Y Y}
% ====== Header (two rows: dataset names + locations) ======
\rowcolor{nmHeader}
\textbf{\color{nmHeadTxt} Characteristic} &
\textbf{\color{nmHeadTxt} Category} &
\textbf{\color{nmHeadTxt} EMBED (N=8,481)} &
\textbf{\color{nmHeadTxt} VinDr (N=5,000)} &
\textbf{\color{nmHeadTxt} RSNA (N=11,913)} \\
\rowcolor{nmHeader}
\textbf{\color{nmHeadTxt}} & \textbf{\color{nmHeadTxt}Location} &
\textcolor{nmHeadTxt}{Emory, GA, USA} &
\textcolor{nmHeadTxt}{Vietnam} &
\textcolor{nmHeadTxt}{USA} \\
\rowcolor{nmHeader}
\textbf{\color{nmHeadTxt}} & \textbf{\color{nmHeadTxt}\#Images} &
\textcolor{nmHeadTxt}{45{,}736} &
\textcolor{nmHeadTxt}{20{,}486} &
\textcolor{nmHeadTxt}{54{,}706} \\

\midrule

% --- Cancer ---
\multirowcell{2}[-.2ex][l]{\cellcolor{nmSubhdr}\textbf{Cancer}} &
\textbf{Present} & \textbf{333 (3.9\%)} & \textbf{481 (9.6\%)} & \textbf{486 (4.1\%)} \\
& Absent & 8{,}148 (96.1\%) & 4{,}519 (90.4\%) & 11{,}427 (95.9\%) \\
\addlinespace[3pt]\midrule

% --- Mass ---
\multirowcell{2}[-.2ex][l]{\cellcolor{nmSubhdr}\textbf{Mass}} &
\textbf{Present} & \textbf{1{,}588 (18.7\%)} & \textbf{584 (11.7\%)} & \NA \\
& Absent & 6{,}893 (81.3\%) & 4{,}416 (88.3\%) & \NA \\
\addlinespace[3pt]\midrule

% --- Calcification ---
\multirowcell{2}[-.2ex][l]{\cellcolor{nmSubhdr}\textbf{Calcification}} &
\textbf{Present} & \textbf{1{,}435 (16.9\%)} & \textbf{220 (4.4\%)} & \NA \\
& Absent & 7{,}046 (83.1\%) & 4{,}780 (95.6\%) & \NA \\
\addlinespace[3pt]\midrule

% --- Architectural Distortion ---
\multirowcell{2}[-.2ex][l]{\cellcolor{nmSubhdr}\textbf{Architectural Distortion}} &
\textbf{Present} & \textbf{183 (2.2\%)} & 65 (1.3\%) & \NA \\
& Absent & 8{,}298 (97.8\%) & 4{,}935 (98.7\%) & \NA \\
\addlinespace[3pt]\midrule

% --- Asymmetry ---
\multirowcell{2}[-.2ex][l]{\cellcolor{nmSubhdr}\textbf{Asymmetry}} &
\textbf{Present} & \textbf{1{,}841 (21.7\%)} & 96 (2\%) & \NA \\
& Absent & 6{,}640 (78.3\%) & 4{,}904 (98.0\%) & \NA \\
\addlinespace[3pt]\midrule

% --- Nipple_Retraction ---
\multirowcell{2}[-.2ex][l]{\cellcolor{nmSubhdr}\textbf{Nipple Retraction}} &
\textbf{Present} & \NA & 22 (0.5\%) & \NA \\
& Absent & \NA & 4{,}978 (99.5\%) & \NA \\
\addlinespace[3pt]\midrule

% --- Skin_Retraction  ---
\multirowcell{2}[-.2ex][l]{\cellcolor{nmSubhdr}\textbf{Skin Retraction }} &
\textbf{Present} & \NA & 13 (0.2\%) & \NA \\
& Absent & \NA & 4{,}987 (99.8\%) & \NA \\
\addlinespace[3pt]\midrule

% --- Skin Thickening   ---
\multirowcell{2}[-.2ex][l]{\cellcolor{nmSubhdr}\textbf{Skin Thickening  }} &
\textbf{Present} & \NA & 34 (0.7\%) & \NA \\
& Absent & \NA & 4{,}966 (99.3\%) & \NA \\
\addlinespace[3pt]\midrule

% --- Suspicious Lymph Node ---
\multirowcell{2}[-.2ex][l]{\cellcolor{nmSubhdr}\textbf{Suspicious Lymph Node }} &
\textbf{Present} & \NA & 53 (1\%) & \NA \\
& Absent & \NA & 4{,}947 (99.0\%) & \NA \\
\addlinespace[3pt]\midrule

% --- BI-RADS ---
\multirowcell{6}[-.2ex][l]{\cellcolor{nmSubhdr}\textbf{BI\mbox{-}RADS}} 
& 0 & 133 (1.6\%) & \NA & 3{,}287 (27.5\%) \\
& 1 & 2{,}117 (25.0\%) & 2{,}515 (50.3\%) & 3{,}151 (26.4\%) \\
& 2 & 3{,}057 (36.0\%) & 1{,}568 (31.4\%) & 604 (5.07\%) \\
& 3 & 1{,}719 (20.3\%) & 436 (8.7\%) & \NA \\
& 4 & 1{,}262 (14.9\%) & 368 (7.4\%) & \NA \\
& 5 & 193 (2.27\%) & 113 (2.3\%) & \NA \\
\addlinespace[3pt]\midrule

% --- Race ---
\multirowcell{6}[-.2ex][l]{\cellcolor{nmSubhdr}\textbf{Race}} 
& African American & 3{,}823 (45.2\%) & \NA & \NA \\
& Alaskan Native & 25 (0.3\%) &\NA& \NA \\
& Asian & 415 (4.9\%) & \NA & \NA \\
& Caucasian or White & 3{,}363 (39.8\%) & \NA & \NA \\
& Multiple & 39 (0.5\%) & \NA & \NA \\
& Native Hawaiian & 108 (1.3\%) &\NA & \NA \\
& Unreported & 680 (8.0\%) & \NA & \NA \\
\addlinespace[3pt]\midrule

% --- Density ---
\multirowcell{4}[-.2ex][l]{\cellcolor{nmSubhdr}\textbf{Density}} &
Almost entirely fatty (A) & 133 (1.6\%) & 24 (0.48\%) & 553 (4.64\%) \\
& Scattered fibroglandular (B) & 2{,}117 (25.0\%) & 469 (9.37\%) & 2{,}511 (21.1\%) \\
& Heterogenously Dense (C) & 3{,}057 (36.0\%) & 3{,}826 (76.2\%) & 2{,}436 (20.4\%) \\
& Extremely Dense (D) & 1{,}719 (20.3\%) & 681 (13.6\%) & 309 (2.59\%) \\

\bottomrule
\end{tabularx}

\begin{tablenotes}[flushleft]\footnotesize
\item \textbf{Note:} \NA{} indicates data not available for that dataset. Bold values denote positive findings.
\end{tablenotes}
\end{threeparttable}
\end{table*}

\textbf{Downstream tasks and datasets.} The primary in-distribution evaluation utilizes a curated subset of the public EMBED dataset~\cite{jeong2023emory} for four classification tasks: cancer, mass, calcification, and architectural distortion. To construct the training and evaluation cohorts, we follow a rigorous case-definition protocol as per~\cite{hwang2023impact}.  The positive cohort is designed to identify screening mammograms that precede a cancer diagnosis. First, all diagnostic mammograms with a BI-RADS assessment of 4, 5, or 6 are identified. For each of these cases, the most recent prior screening mammogram within a 180-day window is selected. The negative cohort consists of all screening studies with a BI-RADS assessment of 1 or 2. Patients appearing in the positive cohort are stringently excluded from the negative cohort to prevent data contamination. For both cohorts, only standard 2D craniocaudal (CC) and mediolateral oblique (MLO) views are included.

We further evaluate the generalization capabilities of Mammo-FM on two public, out-of-distribution datasets. The first is the VinDr dataset~\cite{nguyen2023vindr} (5,000 exams), which facilitates multiple evaluation tasks. These include classification of mass, calcification, and breast density. Following~\cite{kerlikowske2022association, ibrokhimov2022two, yoon2023artificial}, for cancer classification on VinDr, which lacks explicit malignancy labels, patient-level BI-RADS assessments are used as a proxy, where BI-RADS 1-2 are considered normal and BI-RADS 4-5 are considered malignant, following prior work~\cite{ghosh2024ladder, wen2024breast}.
We follow the standard train/validation/test splits provided with the dataset. Our detection evaluation task leverages VinDr dataset to localize mass and calcification. The second OOD dataset is the RSNA Breast Cancer Detection challenge dataset~\cite{rsna-breast-cancer-detection}, which is used to classify cancer. For zero-shot evaluation, we consider the entire VinDr and RSNA dataset as a test set and report the performance. For linear probing and full fine-tuning experiments, we use the standard training–validation–test splits of the VinDr dataset. For the same on RSNA, we construct a custom 70/20/10 split for training, validation, and testing, ensuring strict patient-level separation across subsets.

\subsubsection*{Baselines}
We benchmark the Mammo-FM multi-institutional model's performance by comparing it against a series of state-of-the-art models across all downstream tasks. For zero-shot classification, we compare the multi-institutional model with three vision-language models. The primary ablation baseline is Mammo-FM (UPMC), a variant that follows the identical architecture and pre-training protocol as the main model but uses only the single-institution UPMC dataset. This comparison emphasizes the impact of multi-institutional data. The second baseline is MedSigLIP, the specialized vision encoder from the MedGemma model~\cite{sellergren2025medgemma}, which is a SigLIP model fine-tuned on a large corpus of medical images. For this baseline, input mammograms are resized to its required 448$\times$448 pixel resolution. Third, to evaluate against representations learned from a different radiological domain, we use CXR-CLIP-RN50~\cite{you2023cxr}, a ResNet-50-based vision-language model pre-trained on chest X-ray datasets (handling resolution of $224\times 224$). Finally, for the linear probing and fine-tuning settings, the evaluation incorporates DINOv3~\cite{simeoni2025dinov3}, a state-of-the-art self-supervised vision model, as a strong image-only baseline. For DINOv3, input images are resized to 512$\times$512 pixels. For detection tasks, we compare against two baselines: a RetinaNet detection head built on Mammo-FM (UPMC), and CXR-CLIP-RN50  image encoder. 

\subsubsection*{Metric}
We report AUROC scores for binary classification tasks to predict
mass, calcification, and cancer, and accuracy for multi-class classification tasks to predict density. For localization, we report mean average precision (mAP) with IoU=0.5 and detector
confidence thresholded to 0.05~\cite{lin2017focal} unless specified.

\subsection{Risk prediction with the image-encoder of Mammo-FM}

Accurate and interpretable breast cancer risk prediction is essential for enabling personalized screening and fostering clinical trust in AI-driven diagnostics. The state-of-the-art MIRAI model~\cite{yala2021toward} achieves strong predictive accuracy by aggregating multi-view mammographic representations through a transformer, but its latent features remain opaque and lack clinical interpretability. AsymMIRAI~\cite{donnelly2024asymmirai} advances this direction by introducing an interpretable framework based on bilateral dissimilarity between left–right breast representations. However, both models rely on MIRAI’s original ResNet-18 encoder, which is limited in representational capacity and unaligned with textual semantics.

To address these limitations, we develop two enhanced risk predictors—MIRAI w/ Mammo-FM and AsymMIRAI w/ Mammo-FM -- that integrate Mammo-FM’s multimodally aligned visual encoder trained on paired mammograms and radiology reports. The MIRAI variant leverages a transformer aggregator for multi-view risk estimation, while the AsymMIRAI variant computes localized bilateral dissimilarity for interpretable anomaly detection. Both models are trained with knowledge distillation from MIRAI’s risk outputs, preserving predictive fidelity. Furthermore, by coupling with a Sparse Autoencoder–based interpretability module, our framework provides dual interpretability—linking visual activations in mammograms to their corresponding textual descriptors in radiology reports.

\begin{figure*}[h]
\begin{center}
\includegraphics[width=0.88\textwidth]{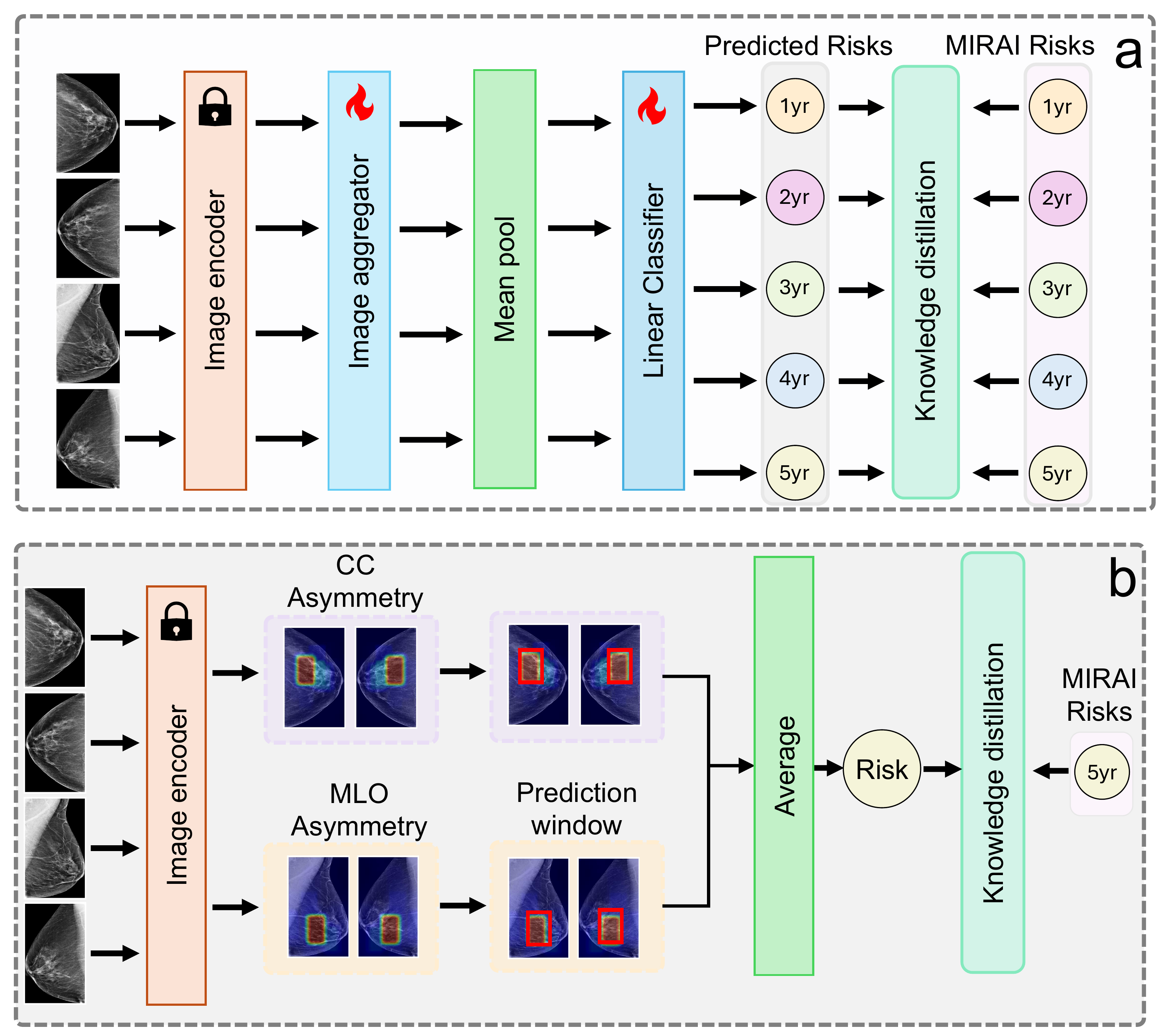}
\caption{\textbf{Integration of Mammo-FM representations into risk prediction frameworks.}
\textbf{a.} MIRAI w/ Mammo-FM. Each of the four standard mammographic views (LCC, LMLO, RCC, RMLO) passes through an independent, frozen Mammo-FM image encoder to produce a 2,048-dimensional feature. A lightweight transformer aggregator fuses view-specific representations enriched with side, view, and time embeddings. The aggregated features are mean-pooled and passed through a linear classifier to generate 1- to 5-year risk logits. The model learns under a knowledge-distillation objective from the original MIRAI teacher, aligning its multi-year risk predictions.
\textbf{b.} Each mammographic view (LCC, LMLO, RCC, RMLO) is encoded using the frozen Mammo-FM image encoder to produce text-aligned visual representations. The model computes localized asymmetry maps between left–right pairs for both CC and MLO views to identify regions of maximal dissimilarity. From these asymmetry maps, the two highest-scoring prediction windows are selected and averaged to derive the final risk estimate. The risk predictor is optimized using knowledge distillation from the original MIRAI’s 5-year risk output, enabling interpretable risk modeling grounded in bilateral tissue differences.}
\label{fig:app_prognosis}
\end{center}
\end{figure*}

\subsubsection *{MIRAI w/ Mammo-FM}
\paragraph{Architecture.}
MIRAI w/ Mammo-FM extends the original MIRAI architecture by replacing its ResNet-18 encoder with a frozen Mammo-FM (multi-institution) encoder that jointly aligns image and text representations. Each of the four standard mammographic views (LCC, RCC, LMLO, RMLO) passes independently through the Mammo-FM image encoder to produce a 2,048-dimensional feature embedding. The model stacks these four embeddings into a sequence and forwards them to a lightweight transformer that integrates contextual metadata before classification. Each 2,048-dimensional feature is linearly projected to a 512-dimensional hidden space. The transformer augments every token with three learned metadata embeddings—time, view, and side—each of 32 dimensions. These embeddings are concatenated into a 96-dimensional vector, which is then linearly mapped to 512 dimensions and added to the image tokens, producing context-aware representations.

The sequence encoder comprises three transformer layers with eight attention heads, a hidden size of 512, and a dropout of 0.25. Each layer includes multi-head self-attention and a two-layer feed-forward block, both wrapped with pre-activation residual connections and layer normalization. This compact configuration supports efficient cross-view communication and long-range contextual reasoning across craniocaudal (CC) and mediolateral oblique (MLO) views, as well as left–right asymmetry. The transformer outputs a four-token sequence, one per view. The model mean-pools across tokens, applies a ReLU activation and dropout, and passes the result through a linear classifier to produce risk logits corresponding to a 1–5-year breast cancer prediction. This structure retains MIRAI’s multi-view risk modeling capabilities while leveraging Mammo-FM’s semantically grounded, multimodal feature space.

\paragraph{Training.}
MIRAI w/ Mammo-FM follows a knowledge distillation (KD) framework, using the original MIRAI model as the teacher. For each screening exam, the teacher produces 1–5-year breast cancer risk logits, which serve as soft supervision targets. The student network, MIRAI w/ Mammo-FM, learns to replicate these temporal risk distributions while leveraging the multimodally aligned Mammo-FM encoders. We only train the transformer aggregator and the linear layer, keeping the Mammo-FM image encoder fixed.

\subsubsection *{AsymMIRAI w/ Mammo-FM}
\paragraph{Architecture.}
Fig.~\ref{fig:app_prognosis}b illustrates the schematic of AsymMIRAI w/ Mammo-FM.
The model architecture follows the design of AsymMIRAI~\cite{donnelly2024asymmirai}, which processes four standard mammographic views -- LCC, RCC, LMLO, and RMLO -- through independent encoders. Each view passes through the shared multi-institutional (Boston University, UPMC, EMBED, and Mayo Clinic) Mammo-FM vision encoder. These encoders remain frozen during risk model training to preserve their pre-learned semantic alignment between images and radiological text. By maintaining shared and frozen visual representations, the model ensures consistency across views while leveraging the generalizable and text-aligned features of the Mammo-FM foundation model.

\paragraph{Training.}
The interpretable risk predictor adapts the architectural framework of AsymMirai~\cite{donnelly2024asymmirai}. The model's four parallel vision encoders are initialized with weights from the pre-trained, multi-institutional Mammo-FM model and remain frozen throughout training. For a given examination, the four standard screening views (LCC, RCC, LMLO, and RMLO) are passed through these encoders to obtain their respective latent representations, \(Z_{\mathrm{VIEW}} \in \mathbb{R}^{C \times H \times W}\), where \(C\) is the channel dimension and \((H, W)\) are the spatial dimensions.

The core of the model is a bilateral dissimilarity module that processes these representations. For each bilateral pair (e.g., LMLO/RMLO), the module first computes a localized bilateral dissimilarity map. This involves taking the absolute, channel-wise difference between the latent tensors:
\begin{equation}
D = \bigl| Z_{\mathrm{right}} - Z_{\mathrm{left}} \bigr| \in \mathbb{R}^{C \times H \times W}.
\end{equation}
A max-pooling operation with a window size of \((a,b)\) is then applied to this difference map, yielding a max-pooled difference map, \(M \in \mathbb{R}^{C \times H' \times W'}\). Each vector \(M_{:,h,w} \in \mathbb{R}^{C}\) in this map summarizes the strongest left-right discrepancies within a local neighborhood, providing robustness to minor misalignments.

Next, a trainable linear layer with weights \(W_{\mathrm{VIEW}} \in \mathbb{R}^{C \times C}\) projects each local vector \(M_{:,h,w}\) to a scalar asymmetry score, creating a 2D asymmetry score map:
\begin{equation}
\Delta_{h,w} = \bigl\| W_{\mathrm{VIEW}}\, M_{:,h,w} \bigr\|_2, \quad \Delta \in \mathbb{R}^{H' \times W'}.
\end{equation}

The single maximal value in this map is identified as the prediction-window score for that view pair:
\begin{equation}
p_{\mathrm{VIEW}} = \max_{\,h,w} \; \Delta_{h,w}.
\end{equation}

Finally, the model generates a single, examination-level risk score by averaging the prediction-window scores from the MLO and CC view pairs:
\begin{equation}
R = \tfrac{1}{2}\,\bigl(p_{\mathrm{MLO}} + p_{\mathrm{CC}}\bigr).
\end{equation}

During training, only the weights of the per-view linear layers (\(W_{\mathrm{MLO}}\) and \(W_{\mathrm{CC}}\)) are trainable. In contrast to the original AsymMirai, which trains directly on binary cancer labels, our model is optimized using a knowledge distillation framework~\cite{hinton2015distilling} from MIRAI's 5-year risk. For each 5-year risk horizon, the model is trained with a composite loss function that combines a standard binary cross-entropy term for direct supervision from ground-truth cancer labels with a knowledge distillation loss. This second term encourages the model to approximate the 5-year risk predictions from MIRAI as a ``teacher'' model. A hyperparameter \(\alpha\) balances these two objectives, training the model to simultaneously mimic MIRAI's predictions and accurately classify true cancer outcomes.

% In contrast to the original AsymMirai, which trains directly on binary cancer labels, our model learns from the nuanced risk predictions of the original MIRAI model through a knowledge distillation framework. For a given risk horizon (\eg 5-year risk), the trainable weights are optimized using a composite loss function. This loss combines a standard Binary Cross-Entropy with Logits term, which provides direct supervision from the ground-truth cancer labels, with a knowledge distillation term. The distillation term is formulated as the Kullback-Leibler (KL) divergence between the temperature-scaled Bernoulli distributions of our model's predictions and the corresponding risk predictions from the MIRAI ``teacher'' model. A hyperparameter \(\alpha\) balances the contribution of these two loss components, training the model to simultaneously mimic MIRAI's predictions and accurately classify true cancer outcomes.

\subsubsection *{Interpreting the risk using SAE}
\paragraph{Matryoshka Sparse Autoencoder Training.}
The interpretability framework relies on a Matryoshka Sparse Autoencoder (SAE) to decompose Mammo-FM's complex representations into a discrete, interpretable vocabulary. We train the SAE on approximately 1.2 million individual spatial patch features from the Boston Medical Center (BMC) dataset (using the same training cohort used for Mammo-FM pretraining). These 2048-dimensional features, extracted from the Mammo-FM image encoder, correspond to a $48 \times 29$ spatial grid. The MSAE architecture maps these features to a 16,384-dimensional sparse latent space (an 8 $\times$ expansion factor).
Training on individual patch features rather than pooled representations is critical for preserving the fine-grained spatial information necessary for localizing findings. The model is trained using a multi-scale reconstruction loss across nested sparsity levels (from $k=64$ to $k=16,384$) to enforce this sparse, hierarchical decomposition. The latent space forms a structured hierarchy where neurons specialize in distinct mammographic patterns \eg masses, calcifications, or architectural distortions. Each neuron behaves as a localized concept unit that preserves both spatial and semantic coherence. The reconstruction objective combines an $\ell_2$ loss with a sparsity-inducing $\ell_1$ penalty on activations, optimized using Adam with a learning rate of 
$1 \times 10^{-4}$.
This formulation produces a compact, interpretable representation that retains the discriminative fidelity of the original Mammo-FM features while enabling direct neuron-level analysis
Before the interpretability analysis, we validate that these MSAE-reconstructed features retain full discriminative fidelity, achieving classification and detection performance nearly identical to the original Mammo-FM features (Fig.~\ref{fig:prognosis}h).

\paragraph{Quantifying Neuron Contribution to risk via Shapley Values.}
To identify the neurons most causally linked to breast cancer risk, we compute Shapley values between MSAE activations and risk predictions. The analysis estimates each neuron’s marginal contribution to the predicted risk by comparing the model’s output with and without that neuron active. Exact computation across all $2^{16,384}$ neuron subsets is infeasible, so we use a permutation-based Monte Carlo approximation.

In each iteration, the algorithm samples a random permutation of all neurons and measures a target neuron’s marginal contribution by evaluating the difference in risk score when the neuron is included versus excluded, given all neurons preceding it in the permutation. Averaging these marginal effects over 100 random permutations yields an unbiased estimate of the neuron’s Shapley value. Positive Shapley values indicate risk-increasing neurons—those that drive higher predicted risk when activated. Ranking these neurons by magnitude provides a causal ordering of the features most responsible for the model’s risk assessment.

\paragraph{Image-Level Interpretability: Spatial Activation Heatmaps.}
We provide image-level interpretability by visualizing the spatial activity of high-risk neurons identified via Shapley analysis. For a given exam, the activation value for a specific neuron is extracted for each patch in the $48 \times 29$ spatial grid. This process generates an activation heatmap that precisely localizes the mammographic regions responsible for the neuron's activation and, consequently, its contribution to the risk score.

\paragraph{Text-Level Interpretability: Causal Language Alignment.} The framework provides text-level interpretability by leveraging Mammo-FM's intrinsic contrastive text-image alignment. To connect a high-risk neuron to a clinical description, we perform a causal ablation experiment. First, the model computes a baseline cosine similarity score between an MSAE-reconstructed image feature and every sentence in the paired radiology report. Next, it intervenes on the single target neuron by setting its activation to zero and generates a new, ``ablated'' reconstructed feature. It then recalculates all image-to-sentence similarities. The sentence that exhibits the largest drop in similarity is selected as the neuron's semantic interpretation, causally linking its visual activation to a specific textual finding.

\subsubsection*{Dataset and Experimental setup}
% We use BU dataset for training and ID evaluation.
% The same train-val-test split used which was used for pretraining Mammo-FM.
% For each exam, we extract the risk for 1yr to 5yr for BU data from MIRAI model. For each yr, we train our model with the MIRAI risk using knowledge distillation.
% For OOD, evaluation, we use the whole RSNA dataset. RSNA dataset has cancer label. We treat 1 yr risk as the predicted cancer cancer label (explain why with proper reasoning.) 
Both the risk predictors are trained and evaluated for in-distribution (ID) performance using the Boston University (BU) dataset. To ensure consistency and prevent data leakage, the evaluation utilizes the same patient-level train, validation, and test splits leveraged during the Mammo-FM pre-training phase. For the knowledge distillation training process, we extract the 1- to 5-year risk predictions for each examination in the BU dataset from the original, publicly available MIRAI model. These extracted MIRAI risk scores serve as the ``teacher'' labels for training the model on each respective risk horizon.

We conduct the out-of-distribution (OOD) evaluation on the RSNA and VinDr datasets to estimate the model's generalization capabilities. Because these datasets contain only binary cancer labels and lack multi-year outcome data, we evaluate the model’s 1-year predicted risk as a proxy for cancer detection.
This experimental design is chosen because a 1-year risk prediction task is functionally analogous to a short-term cancer detection task. 
Prior work~\cite{donnelly2024asymmirai} has shown that MIRAI’s 1-year risk score aligns closely with short-term cancer detection tasks, supporting its use as a surrogate for cancer labeling in this setting. We optimize only the view-specific linear layers while keeping the shared Mammo-FM encoders frozen. We use a learning rate of 0.005 and a mini-batch size of 30 across all experiments.

\subsubsection*{Baselines and Evaluation setup}
We benchmark against two established models: MIRAI and AsymMIRAI. On the BU screening cohort (in-distribution), we extract MIRAI’s and AsymMIRAI’s risk estimates for 1– to 5–year horizons and evaluate examination-level AUROC at each horizon. The Mammo-FM risk predictor reports the same set of horizon-specific AUROCs for a direct comparison. For out-of-distribution evaluation, we use the RSNA and VinDr dataset with binary cancer labels. MIRAI, AsymMIRAI, and the Mammo-FM-based risk predictors each produce a 1-year risk, which we treat as the short-horizon proxy for a cancer/no-cancer decision on RSNA and VinDr. We compute patient-level AUROC on the full RSNA cohort to assess generalization.

\subsection{Mammo-GRG}

The Mammo Grounded Report Generator (Mammo-GRG) generates clinically accurate and grounded radiology reports from multi-view screening mammograms. It is a multi-modal model composed of a frozen vision encoder, a multimodal projector, and a large language model. Mammo-GRG follows a LLaVA-style~\cite{liu2023visual, li2023llava} two-stage training protocol designed to align the visual and language modalities and then fine-tune for instruction-following. Both stages optimize a standard next-token prediction loss.

Stage 1 aligns the image features with the LLM by training the multimodal projector. In this stage, both the vision encoder and the LLM remain frozen. The training objective for this stage is to predict the full ground-truth radiology report from the image features, effectively teaching the projector to translate visual information into a format the LLM can comprehend.
Stage 2 follows an end-to-end instruction tuning where both the multimodal projector and the LLM are fine-tuned. This stage utilizes a rich dataset of question-answer pairs generated from the original mammography reports, training the model to predict answers to specific instructions and generate more detailed, conversational outputs beyond simple report generation. Refer to Tab.~\ref{tab:mammo_grg_system_prompt} for the system prompt to create the preliminary report after stage 2.

Mammo-GRG aims to solve three primary challenges. The architecture and training are designed to address three primary challenges. First, for multi-view image integration, the model processes the four mammogram views (LCC, LMLO, RCC, RMLO) through four parallel, weight-shared vision encoders. The resulting image features are enclosed within unique tokens (\texttt{\textless LCC\textgreater, \textless LMLO\textgreater}, \etc) to inform the LLM of each view's identity. To distinguish these visual tokens, we augment each view's projected queries with a learned, per-view positional embedding, which functions as a view identifier. Second, to mitigate data imbalance in the training cohorts, we parse the BU and UPMC reports using GPT-4o to create structured labels for mass, calcification, architectural distortion, and asymmetry. We then use these labels to create a case-level binary indicator for finding presence, which enables balanced dataset sampling during training. Third, to ensure the clinical fidelity of the final report, the Mammo-GRG implements a grounding step. Following Stage 2, the instruction-tuned generator produces a preliminary report. In parallel, the core Mammo-FM model performs zero-shot classification on the four input views to extract structured predictions for key clinical findings. In the final stage, an LLM uses both this preliminary report and the structured findings as context to synthesize the two streams of information and generate the final, clinically grounded report. Tab.~\ref{tab:mammo_grg_grounding_prompt} for the system prompt to create the final report from the preliminary report after this grounding stage.

\subsubsection*{Training Data Curation for Mammo-GRG}
We construct the training dataset for Mammo-GRG from the Mammo-FM training splits of the Boston University (BU) and UPMC datasets. From the BU training set, we include all 10,921 examinations from 4,950 patients, where each examination comprises four standard views and a single corresponding report. From the UPMC training set, we select only those examinations that contain all four standard views, resulting in a cohort of 8,890 examinations. For both datasets, we pre-process the raw radiology reports to select only the \texttt{Impression} and \texttt{Findings} sections, discarding all other text.

To create a rich instruction-tuning dataset for Stage 2, we use these cleaned reports as a foundation to generate Mammo-Instruct, a large-scale set of question-answer (QA) pairs using GPT-4o. Mammo-Instuct comprises a total of 359,746 high-quality QA pairs, including 134,895 from the UPMC reports and 163,390 from the BU reports. Mammo-instruct consists of multiple-choice, free-response, description, and conversation-style questions designed to test the model's ability to recall specific facts from the report, and long-answer questions, which require the model to synthesize information and provide more descriptive explanations. Refer to Tab.~\ref{tab:prompt_mammo_instruct} and Tab.~\ref{tab:prompt_multitype} for the prompts to create the Mammo-Instruct dataset.

\subsubsection*{Architecture}
% vision encoder: Mammo-FM vision encoder --  efficient net B5 trained with multiinstitution data.
% textencoder: Llama 3.1 8B   we use
% the 8B Meta Llama 3.1 which includes 32 transformer layers, 32 attention heads, an embedding
% dimension of 4,096, an FFN dimension of 14,336, and a context size of up to 128k tokens

% The multimodal projector, based on the approach described in [14],
% consists of an attention-pooling layer followed by a 2-layer multi-layer perceptron (MLP). The attentionpooling mechanism uses 256 learned latent queries in conjunction with multi-headed cross-attention to
% compress the final layer’s features from the encoder backbone into a fixed-length sequence of image
% tokens. This approach optimizes both training and inference, ensuring that the sequence length remains
% within the LLM’s context window, and making the model focus more on the important features with the
% attention mechanism. The MLP that follows is adapted from LLava 1.6 [26] and includes a single hidden
% layer activated by GeLU, transforming the image tokens to match the LLM’s embedding dimension.

The Mammo-GRG architecture consists of three core components: a vision encoder, a large language model (LLM), and a multimodal projector that aligns them. The vision encoder is the pre-trained, multi-institution Mammo-FM vision encoder with an EfficientNet-B5 backbone. The language model is the Llama 3.1 LLM with 8B parameters, which includes 32 transformer layers, 32 attention heads, an embedding dimension of 4,096, a feed-forward network dimension of 14,336, and a context size of up to 128k tokens.

The multimodal projector, based on the approach described in~\cite{hamamci2024developing}, consists of an attention-pooling layer followed by a 2-layer multilayer perceptron (MLP). The attention-pooling mechanism uses 256 learned latent queries in conjunction with multi-headed cross-attention to compress the final layer’s features from the encoder backbone into a fixed-length sequence of image tokens. This attention-based approach allows the model to focus on the most salient visual features while ensuring the resulting sequence length remains within the LLM’s context window. The subsequent MLP, adapted from LLaVA 1.6~\cite{liu2023visual}, includes a single hidden layer with a GeLU activation function and transforms the image tokens to match the LLM’s embedding dimension.

\subsubsection*{Training}

The development of Mammo-GRG follows a two-stage training protocol that optimizes a standard next-token prediction loss. To manage multi-view image inputs and multi-task text outputs, the tokenizer is augmented with special tokens: view-specific tokens (\textless LCC\textgreater, \textless LMLO\textgreater, etc.) to delineate image inputs, and task-specific tokens (\textless long answer\textgreater, \textless report generation\textgreater, etc.) to guide the model’s output mode.

The first stage is a medical concept alignment step~\cite{li2023llava}. Here, the LLM is frozen, and only the multimodal projector is trained. The model receives the prompt ``Please provide the radiology report for the following 2D screening mammogram <image>'' and is trained to generate the ground-truth report, which aligns the projector with biomedical concepts. The second stage is a medical instruction tuning step. This stage utilizes the full Mammo-Instruct QA dataset to fine-tune the multimodal projector and the LLM. This process fine-tunes the multimodal projector and employs Low-Rank Adaptation (LoRA)~\cite{hu2022lora} to efficiently update the LLM. The LoRA configuration uses a rank of 128 and an alpha of 256. The training follows the same hyperparameter settings as CT-CHAT~\cite{hamamci2024developing}. Across both stages, we freeze the parameters of the Mammo-FM vision encoder. 

\subsubsection*{Baselines}

\textbf{Original Generalist Models.} We benchmark Mammo-GRG against two state-of-the-art generalist medical vision-language models: LLaVA-Med~\cite{li2023llava} and MedGemma~\cite{sellergren2025medgemma}.
LLaVA-Med’s training follows a two-stage curriculum. The first stage, Biomedical Concept Alignment, utilizes the extensive PMC-15M dataset to map biomedical visual concepts to the language model’s embedding space. The second stage, Biomedical Instruction-Tuning, employs a more curated dataset derived from five common imaging modalities—including computed tomography, magnetic resonance imaging, and histopathology—to refine the model's conversational and instruction-following capabilities. For LLaVA-Med baseline, we utilize  \texttt{llava-med-v1.5-mistral-7b} variant. It integrates a general-purpose CLIP vision transformer (ViT-L/14) with the Mistral-7B large language model. We resize the mammograms to 224 $\times$  224 pixels as an input to the image encoder of LLaVA-Med. 
The second baseline is the instruction-tuned MedGemma-27B model (\texttt{medgemma-27b-it}), a foundation model specifically optimized for medical text and image comprehension. This 4-billion-parameter model combines the Gemma 3 language architecture with a specialized vision encoder, MedSigLIP.  Unlike the general-purpose encoder in LLaVA-Med, MedSigLIP originates from the SigLIP-400M model and undergoes extensive fine-tuning on over 33 million medical image-text pairs. The language component of MedGemma benefits from a diverse training corpus that spans both multimodal and text-only medical data. As these baselines accept only a single image, we generate a report for each of the four views independently and concatenate the results. We resize the mammograms to 224$\times$224 and 448$\times$448 pixels as an input to the image encoder of LLaVA-Med and Medgemma, respectively. 
\textbf{Hybrid Encoder Baselines.} To investigate the contribution of Mammo-GRG’s vision encoder, we develop two additional hybrid baselines, replacing its encoder with those from LLaVA-Med and MedGemma. 
Specifically, these baselines replace the Mammo-FM vision encoder with either the LLaVA-Med's ViT-L/14 encoder (with an input resolution of 224$\times$224) or MedGemma's MedSigLIP encoder (with an input resolution of 448$\times$448). Also, these hybrid baselines adopt the same multi-view architecture as Mammo-GRG, employing four parallel instances of the respective generalist vision encoder to process the four mammogram views.
Further, both of these baselines retain the same multimodal projector and Llama-3.1-8B large language model used in Mammo-GRG. They follow the identical two-stage training protocol, including feature alignment and instruction tuning on our Mammo-Instruct dataset, to ensure a direct and fair comparison of the vision encoders' capabilities.

\subsubsection*{Evaluation setup}

The evaluation of generated mammography reports employs a stringent, multi-faceted methodology to assess lexical similarity, clinical correctness, and pathology-specific detection accuracy.
\textbf{Lexical Similarity Metrics.}
First, we compute standard natural language processing (NLP) metrics to measure lexical overlap between the generated and ground-truth reports. These include BLEU-1, which assesses unigram precision, and ROUGE-L, which measures the longest common subsequence to evaluate fluency and recall.

\textbf{Clinical Correctness using an LLM-based Metric.}
Recognizing the limitations of lexical metrics in capturing clinical accuracy, the primary evaluation utilizes a framework inspired by the GREEN metric~\cite{ostmeier2024green}. We adapt this approach for mammography by using an LLM judge to perform a structured comparison between the candidate (generated) and reference (ground-truth) reports. In all our experiments, we use GPT-4o-mini as the LLM judge. The LLM judge assesses each report pair against five mammography-specific error categories: 1) \textbf{False report of a finding:} The candidate report mentions a finding (e.g., mass, calcification, asymmetry, architectural distortion) that is not in the reference; 2) \textbf{Missing a finding:} The candidate report omits a finding mentioned in the reference; 3) \textbf{Mischaracterization of a finding:} A finding is present in both reports, but its characteristics (e.g., size, margins, stability) are described incorrectly; 4) \textbf{Misidentification of location/laterality:} A finding is correctly identified, but its location (e.g., quadrant, depth) or laterality is wrong; \textbf{Incorrect BI-RADS score:} The final BI-RADS assessment differs between reports. 

For each report pair, the LLM judge outputs a structured JSON object containing the counts of matched findings, the counts for each of the five significant error types, and the count of any insignificant stylistic errors. From this structured output, we calculate the final GREEN score. The score's formulation rewards correctly identified findings while penalizing clinically significant errors. Let \#matched findings denote the count of findings correctly identified in the candidate report, and let \#error$_{\text{sig},i}$ represent the count of errors for each of the five predefined significant error categories, $i\in \{(a),...,(e)\}$. The GREEN score is then expressed as:
\begin{equation}
\text{GREEN} = \frac{\#\text{matched findings}}{\#\text{matched findings} + \sum_{i=(a)}^{(e)} \#\text{error}_{\text{sig}, i}}
\end{equation}
The score is bounded between 0 and 1, with higher values indicating greater clinical accuracy.
Refer to Tab.~\ref{tab:green_metric_prompt} for the prompt to compute the GREEN metric for each predicted report.

\section{Results}
\label{sec:results}

\subsection{Multi-Institutional Pretraining Dataset}
\label{sec:pretraining_dataset}
We extract a large, multi-institutional corpus of 2D screening mammograms from four US institutes to pretrain Mammo-FM. The dataset includes 96,567 patients from Mayo Clinic, 13,829 patients from UPMC, 9,900 patients from Boston University (BU), and 20,381 patients from the public EMBED cohort (see Tab.~\ref{tab:main_dataset_char} and~\ref{tab:bu_dataset_characteristics}; the Dataset section in Methods). All sources contain BI-RADS 0–2 exams with CC and MLO views only. We partition all datasets at the patient level into training, validation, and test splits to ensure no patient appears in more than one split. The combined corpus totals 140,677 patients and 821,326 images. Mayo Clinic, UPMC, BU and EMBED contain 465,078, 46,433, 54,776, and 255,039 images respectively.

\subsection{Diagnostic Robustness of Mammo-FM Across ID and OOD Settings}

Mammo-FM aims to capture robust, high-fidelity image representations designed to generalize across unseen out-of-distribution (OOD) settings.
To validate this, we design a rigorous evaluation setup for in-distribution (ID) and out-of-distribution (OOD) settings across five comprehensive tasks: zero-shot diagnosis, linear probing, full fine-tuning, data efficiency, and pathology localization. Zero-shot evaluation evaluates whether the Mammo-FM diagnoses findings without any labeled data or task-specific training, required for deployment at sites with no annotations. Linear probing and full fine-tuning rigorously quantify the quality of the learned image representations when site-specific training data are available. Data efficiency measures the performance of Mammo-FM with very limited training data to verify the feasibility for institutions with sparse training examples. Localization tasks verify precision in pathology detection, a prerequisite for downstream CAD integration. The ID evaluation leverages the public EMBED dataset to classify mass, suspicious calcification, architectural distortion, and cancer. For OOD evaluation, we utilize the public VinDr (to classify mass, density, calcification, and cancer) and RSNA (to classify cancer) datasets, as both of them were excluded from pretraining (Tab.~\ref{tab:dataset_eval_characteristics}). We benchmark Mammo-FM against multiple baselines, including a UPMC-only variant, generalist FMs (\eg MedSigLIP the vision encoder of MedGemma), chest-X-ray models (\eg CXR-CLIP-RN50), and self-supervised models (\eg DINOv3). We report classification and detection performance using AUROC and mAP, respectively.

\begin{figure}[p]
    \centering
    \includegraphics[width=0.81\textwidth]{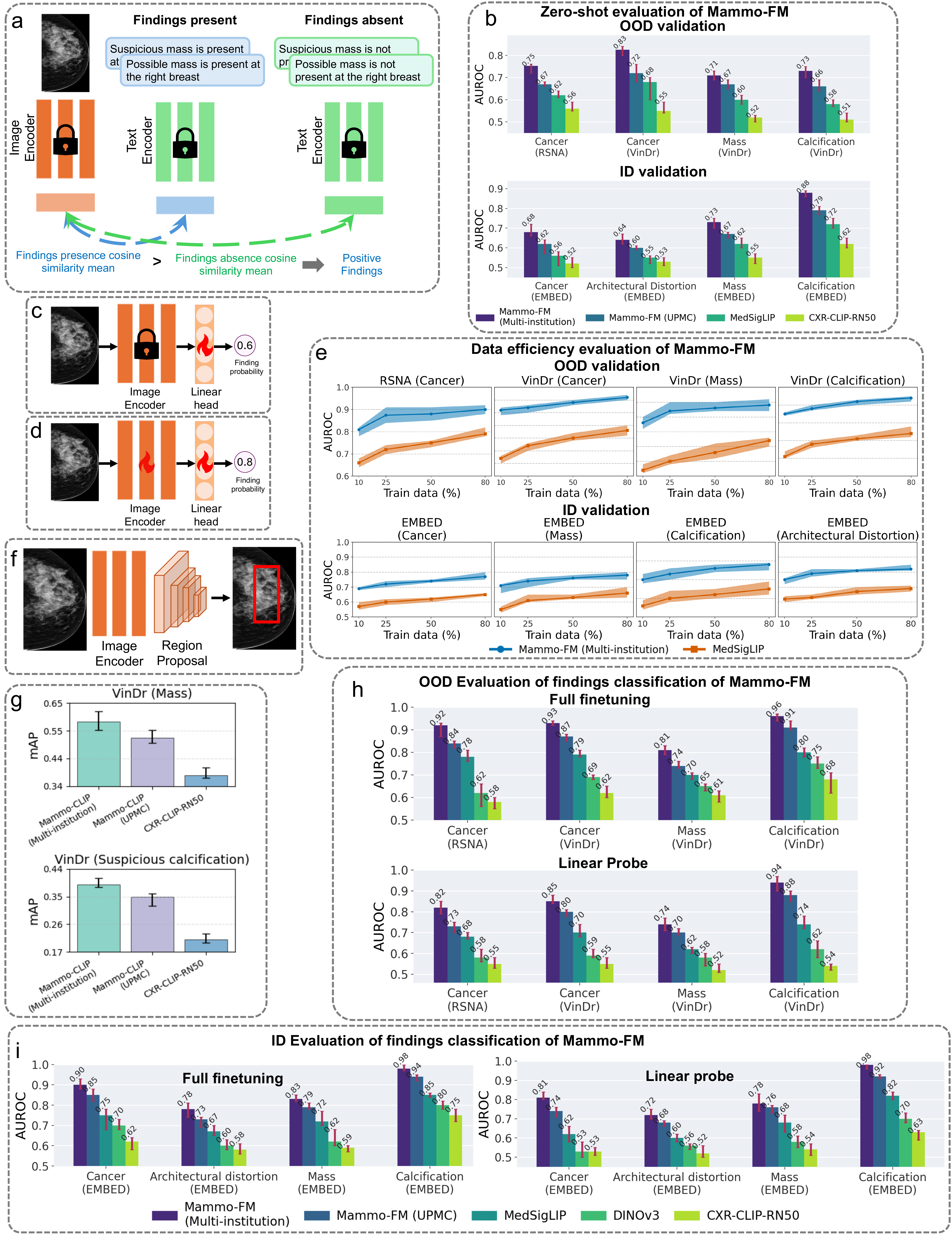}
    \vspace{-0.8em}
    \caption{\textbf{Evaluation of Mammo-FM representations for diagnosing mammographic findings.} \textbf{a.} Schematic overview of zero-shot classification, where aligned image and text encoders predict the presence of mammographic findings (\eg mass, calcification) by measuring image–text embedding similarity, without task-specific training. \textbf{b.} Zero-shot evaluation of breast cancer and finding classification performance on in-distribution (ID; EMBED) and out-of-distribution (OOD; VinDr, RSNA) datasets. \textbf{c.} Illustration of the linear probing setup, in which only a classification head is trained atop the frozen vision encoder to assess representation quality. \textbf{d.} Schematic of the full fine-tuning configuration, where the entire model is optimized end-to-end for downstream diagnostic tasks. \textbf{e.} Data efficiency analysis across ID and OOD datasets, showing progressive improvement in classification performance with increasing fractions of training data. \textbf{f.} Pathology localization using a RetinaNet detection framework with a Mammo-FM vision encoder as backbone. \textbf{g.} Pathology localization performance detecting mass and calcification from VinDr dataset. \textbf{h-i}, Quantitative results for OOD and ID classification under full fine-tuning (top panels) and linear probing (bottom panels), reported as AUROC. Across all tasks and datasets, Mammo-FM (multi-institution) shows higher diagnostic accuracy, superior generalization, and data efficiency compared with single-institution, generalist, and self-supervised baselines -- highlighting the importance of domain-specific, multi-institutional pre-training for clinically robust mammographic representation learning. In all the evaluations, we denote suspicious calcification as calcification for brevity.}  
    \label{fig:diagnosis}
\end{figure}

Mammo-FM (multi-institution) outperforms all baselines across ID and OOD  classification across all four evaluation settings. In training-free zero-shot evaluation, we compute the image-text similarity using the corresponding encoders without task-specific training (Fig.~\ref{fig:diagnosis}a, Methods). Mammo-FM is evaluated on clinically grounded text prompts (Fig.~\ref{fig:app_image_label_augment}, Methods) by converting each finding label (\eg mass) into descriptive sentences developed with a board-certified radiologist, incorporating finding subtype (\eg suspicious, obscured), laterality (left, right), depth (anterior, mid, posterior), and positional descriptors (\eg upper, lower). In zero-shot classification setup, Mammo-FM achieves 13–38\% higher AUROC than MedSigLIP across tasks ($P<0.001$ for all; Fig.~\ref{fig:diagnosis}b). Specifically, for zero-shot cancer classification, it achieves AUROCs of 0.75 (95\% CI: 0.72–0.76) on RSNA and 0.83 (95\% CI: 0.80–0.84) on VinDr, yielding $\sim25$\% and $\sim22$\% relative improvements over MedSigLIP. The multi-institutional pretraining is critical; compared with its UPMC-only variant, the multi-institutional model demonstrates 13–15\% higher AUROC on OOD zero-shot cancer classification for both the datasets. 

To perform the linear probe, we train a linear classifier attached to the frozen vision encoder (Fig.~\ref{fig:diagnosis}c).  In this setting, Mammo-FM (multi-institution) shows strong OOD generalization across the different tasks (Fig.~\ref{fig:diagnosis}h bottom and Fig.~\ref{fig:diagnosis}i right). Specifically, for OOD calcification classification on VinDr, Mammo-FM (multi-institution) achieves an AUROC of 0.94 (95\% CI: 0.90–0.97), representing a 27\% and 11\% improvement over MedSigLIP (AUROC 0.74, 95\% CI: 0.72-0.78; $P = 0.003$) and the single-institution UPMC variant (AUROC 0.88, 95\% CI: 0.85–0.90; $P = 0.003$), respectively. To perform the full-finetuning, we optimize the entire classifier, both the Mammo-FM vision encoder and classifier (Fig.~\ref{fig:diagnosis}d). This setup further reinforces the multi-institutional model's superior performance (Fig.~\ref{fig:diagnosis}h top and Fig.~\ref{fig:diagnosis}i left panel). For OOD cancer classification on VinDr, Mammo-FM (multi-institution) achieves an AUROC of 0.93 (95\% CI: 0.92–0.94). This denotes an 18\% improvement over MedSigLIP (AUROC 0.79, 95\% CI: 0.78–0.81; $P = 0.003$) and a 34\% improvement over the UPMC variant (AUROC 0.69, 95\% CI: 0.68–0.70; $P = 0.003$). Mammo-FM's performance on cancer classification is notable, as it is pretrained on screening mammograms, where reports do not explicitly mention cancer or malignancy. 
Refer to Fig.~\ref{fig:app_density} for the similar results of breast density classification.   

% A similar trend follows for other tasks across EMBED, VinDr and RSNA datasets. This superior performance demonstrates the multi-institutional model's robust generalization to unseen populations, such as the Vietnamese cohort in VinDr. 
To quantify data efficiency, we use varying training data to 10\%, 25\%, 50\%, and 80\%. Cancer classification on VinDr and RSNA uses full fine-tuning; all other findings use linear probing.
Fig.~\ref{fig:diagnosis}e shows Mammo-FM (multi-institution) achieves superior data efficiency on the evaluation settings with limited training data. Specifically, Mammo-FM (multi-institution) trained on only 25\% of the data outperforms MedSigLIP trained on 80\%, delivering a 10–15\% higher AUROC on average across both ID and OOD settings. 
For localization, we attach the Mammo-FM vision encoder as the backbone of a RetinaNet detector and fine-tune end-to-end. Fig.~\ref{fig:diagnosis}g reports the results of pathology localization tasks. For mass detection, the fine-tuned multi-institution Mammo-FM achieves a mAP of 0.58 (95\% CI: 0.55–0.62), a 12\% ($P = 0.005$) improvement over the UPMC variant (mAP = 0.52; 95\% CI: 0.50–0.55). We restrict comparisons to CNN-based models, as the localization pipeline leverages a RetinaNet backbone. 

These results establish Mammo-FM (multi-institution) as a comprehensive diagnostic foundation model. It consistently outperforms generalist and single-institution baselines
across ID and OOD public benchmarks across all four evaluation settings. These findings show that large-scale, domain-specific, multi-institutional pretraining is essential for robust, transferable clinical vision-language models for mammography.
\label{subsec:diagnosis}

\subsection{Interpretable Breast Cancer Risk Prediction with Mammo-FM}
\begin{figure}[p]
    \centering
    \includegraphics[width=0.8\textwidth]{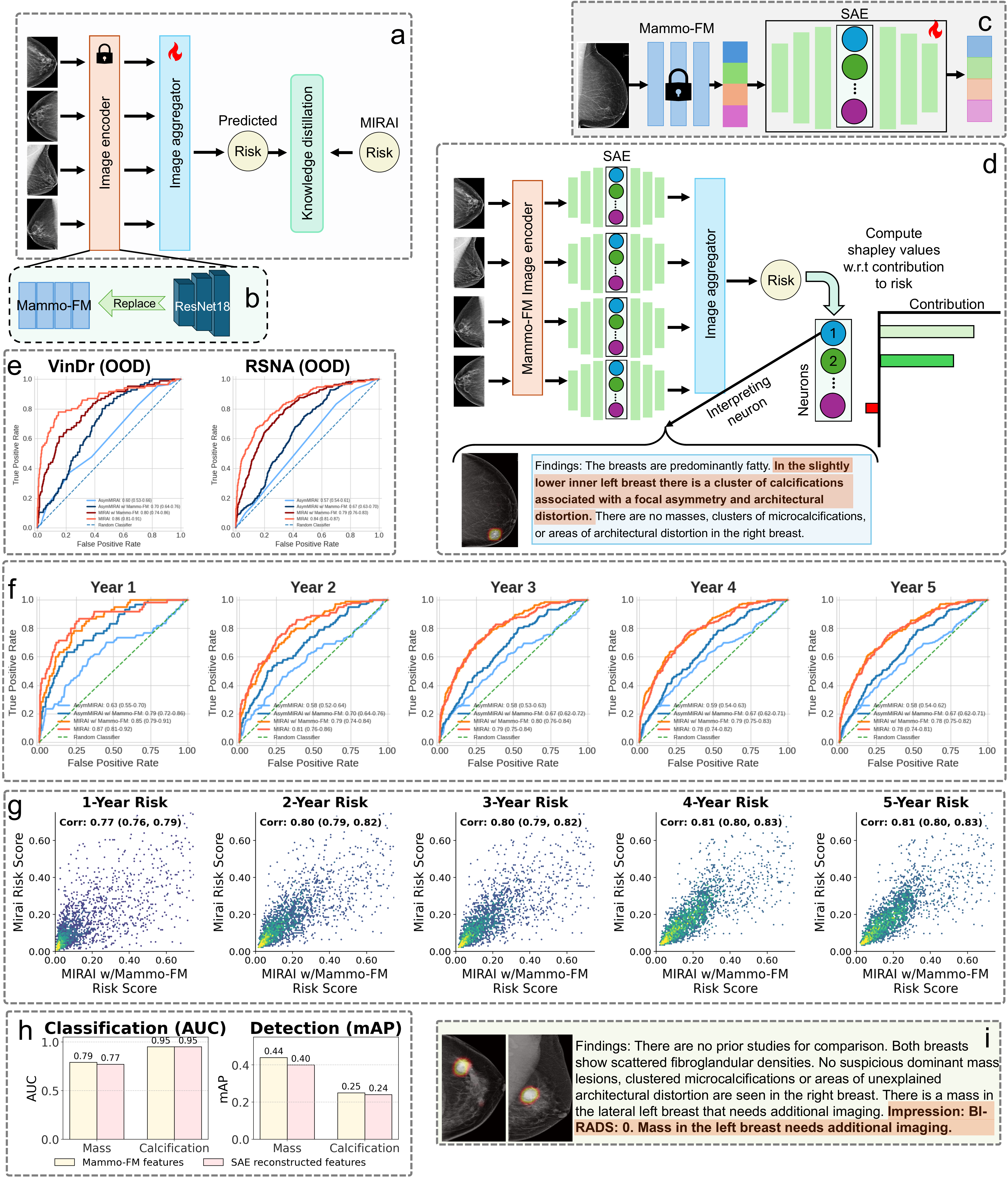}
    \caption{\textbf{Integrating Mammo-FM representations with risk prediction pipelines and interpretable modeling.} \textbf{a.} Training pipeline of the risk predictors -- MIRAI w/ Mammo-FM and AsymMIRAI w/ Mammo-FM using knowledge distillation from 1–5-year risk outputs of the original MIRAI model. We freeze the image encoder throughout the training.
    \textbf{b.} We replace the standard ResNet-18 encoder in MIRAI and AsymMIRAI with a Mammo-FM-trained encoder that aligns images and reports, enabling richer and more transferable representations.
    \textbf{c.} The Matryoshka Sparse Autoencoder (SAE) trains on 2048-dimensional spatial Mammo-FM features (48×29 patch grid) and expands them into a 16,384-dimensional sparse latent space, creating interpretable neuron-level representations. 
    \textbf{d.} Risk predictors use SAE features for interpretability: turning neurons on and off quantifies their effect on risk. Shapley analysis ranks neurons by their contribution, linking top activations to both spatial image regions and key radiology report sentences.
    \textbf{e.} Comparison of AUROC for cancer prediction  MIRAI w/ Mammo-FM, AsymMIRAI w/ Mammo-FM, and their baselines for VinDr (OOD) and RSNA (OOD) datasets. Each curve reports the mean AUROC with 95\% confidence intervals (CIs).
    \textbf{f.} Comparison of AUROC for 1–5-year risk prediction  MIRAI w/ Mammo-FM, AsymMIRAI w/ Mammo-FM, and their baselines using BU (ID) dataset, showing that Mammo-FM-based encoders improve performance while maintaining interpretability. Each curve reports the mean AUROC with 95\% confidence intervals (CIs).
    \textbf{g.} Correlation between MIRAI and MIRAI w/ Mammo-FM risks demonstrates consistent risks across 5-year time horizons using BU (ID) dataset. Reported correlation coefficients (Pearson’s $r$) with 95\% CIs.
     \textbf{h.} SAE-reconstructed features retain discriminative fidelity in classification and detection relative to original Mammo-FM features.
    \textbf{i.} Qualitative visualization of MIRAI w/ Mammo-FM risk interpretation for a high-risk 1-year case (MIRAI risk = 0.74) in the BU dataset shows dual interpretability: image-based saliency highlights a neuron selective for mass features, while the corresponding report sentence yields the highest similarity drop for risk, indicating language-based interpretability. 
    }  
    \label{fig:prognosis}
\end{figure}

% Accurate and interpretable breast cancer risk prediction is essential for personalized screening and for enhancing clinicians' trust. Unlike the state-of-the-art breast cancer risk predictors, Mammo-FM’s vision–language alignment enables both image-level and text-level interpretability for any off-the-shelf risk predictor. 
Mammo-FM’s vision–language alignment grounds risk scores from any breast cancer risk predictor to sentences in the corresponding radiology report, enabling text-level interpretability in prognosis for the first time.
Text-level interpretability is essential because clinicians reason in language and rely on clinical descriptors to trust and audit model behavior.
To investigate this, we utilize two state-of-the-art breast cancer risk models, MIRAI~\cite{yala2021toward} and AsymMIRAI~\cite{donnelly2024asymmirai}. We integrate Mammo-FM (multi-instituition)'s frozen image encoder into both pipelines, yielding MIRAI w/ Mammo-FM and AsymMIRAI w/ Mammo-FM (Fig.~\ref{fig:prognosis}a–b; Methods). We then train both models using MIRAI’s risk outputs through knowledge distillation.  Refer to Fig.~\ref{fig:app_prognosis} for the schematic of the training process of both the Mammo-FM-based risk predictors. To achieve interpretability, we train a Matryoshka Sparse Autoencoder (SAE) on Mammo-FM’s 2048-dimensional image representations (Fig.~\ref{fig:prognosis}c; Methods). The SAE expands these features into a sparse latent space and constructs a neuron-level vocabulary of mammographic patterns. We use Shapley value analysis (Fig.~\ref{fig:prognosis}d; Methods) to quantify each neuron’s causal influence on predicted risk and to identify the strongest risk-increasing neurons. We then provide image-level interpretability by generating spatial activation heatmaps for these neurons, which localize the specific regions that drive the risk score. Mammo-FM’s vision–language alignment further enables text-level explanations, allowing each high-risk neuron to map back to its most relevant radiology-report sentence (See Methods).

For evaluation, we compare the Mammo-FM-based risk predictors with MIRAI and AsymMIRAI as baselines. We first examine the effect of Mammo-FM's image encoder on MIRAI and AsymMIRAI pipelines across ID and OOD setting. So, we train both the Mammo-FM-based risk predictors using the BU train-set. For ID setting, we evaluate 1–5 year risk prediction on the BU test set. To evaluate robustness using OOD datasets -- RSNA and VinDr, we use the 1-year risk score to predict cancer labels. The public RSNA and VinDr datasets provide only binary cancer labels rather than 1–5 year outcomes. Our OOD evaluation therefore, leverages the 1-year risk score as a proxy for existing disease to predict these cancer labels. We use AUROC as the evaluation metric. For interpretability, we test if the SAE-reconstructed representations preserve Mammo-FM’s discriminative features. To do so, we compare the performance of SAE-reconstructed representations with the original Mammo-FM to classify and detect mass and calcification on the OOD VinDr dataset.

% To achieve interpretability, we train a Matryoshka Sparse Autoencoder (SAE) on Mammo-FM's 2048-dimensional image representations (Fig.~\ref{fig:prognosis}c). The SAE expands these features into a 16,384-dimensional sparse latent space, creating an interpretable, neuron-level ``vocabulary'' for mammographic features~\cite{nakka2025mammo}. As exact computation is infeasible, we apply a permutation-based Monte Carlo approximation of Shapley value analysis (Fig.~\ref{fig:prognosis}d) to quantify each neuron’s influence on predicted risk. Specifically, we average a neuron's marginal contribution to the risk score over 100 random permutations, yielding an unbiased Shapley value. 

% We then identify risk-increasing neurons (positive Shapley values) and generate spatial activation heatmaps for high-ranking neurons to localize risk-driving regions. Unlike MIRAI or AsymMIRAI, our framework provides text-level interpretability due to Mammo-FM’s image–text alignment. For each high-risk neuron, we perform causal ablation by suppressing its activation to generate an ablated image representation. We then compute the similarity between this ablated feature and all sentences in the paired radiology report. The sentence with the largest drop in similarity becomes the neuron’s semantic descriptor, directly linking a risk-driving visual concept (\eg a mass-selective neuron) to its corresponding clinical language (\eg ``a suspicious mass is seen'').

Fig~\ref{fig:prognosis}e-f compares the performance of different risk models for the OOD (VinDr and RSNA) and ID (BU) datasets, respectively. For ID evaluation, AsymMIRAI w/ Mammo-FM improves AUROC by $\sim$26–36\% over the original AsymMIRAI across all years. MIRAI w/ Mammo-FM achieves comparable performance to the original black-box MIRAI, with only a minimal AUROC reduction (0.85, 95\% CI: 0.79–0.91) relative to MIRAI’s 0.87 (95\% CI: 0.81–0.92), demonstrating that the multimodal encoder preserves MIRAI’s predictive strength while adding interpretability.
Across both OOD datasets, MIRAI w/ Mammo-FM achieves an AUROC of 0.80 (VinDr) and 0.79 (RSNA), remaining within 3–4\% of the original MIRAI (0.83 and 0.80, respectively).
This consistency under domain shift demonstrates strong OOD robustness.
The MSAE-reconstructed features preserve the discriminative performance (Fig~\ref{fig:prognosis}h) of the original Mammo-FM representations across classification and detection tasks (\eg mass AUROC 0.77 vs. 0.79; calcification 0.95 vs. 0.95; mAP differences $\leq$0.04), indicating minimal information loss. 
This high fidelity enables the use of reconstructed features for neuron-level, multimodal interpretability.
Fig.~\ref{fig:prognosis}i illustrates the qualitative interpretability of the MIRAI w/ Mammo-FM risk predictor for a high-risk 1-year case (MIRAI risk: 0.74). A Shapley-derived saliency heatmap localizes the highest-contributing neuron's activation to a mass in the left breast. The corresponding radiology-report sentence -- ``Mass in the left breast needs additional imaging'' -- shows the largest drop in image–text similarity after neuron ablation, confirming that the neuron encodes a clinically meaningful finding. Refer to Fig~\ref{fig:app_text_interpretability} for more examples.

\label{subsec:prognosis}

\subsection{Multi-View Mammography Reporting with Mammo-GRG}
\begin{figure}[p]
    \centering
    \includegraphics[width=0.85\textwidth]{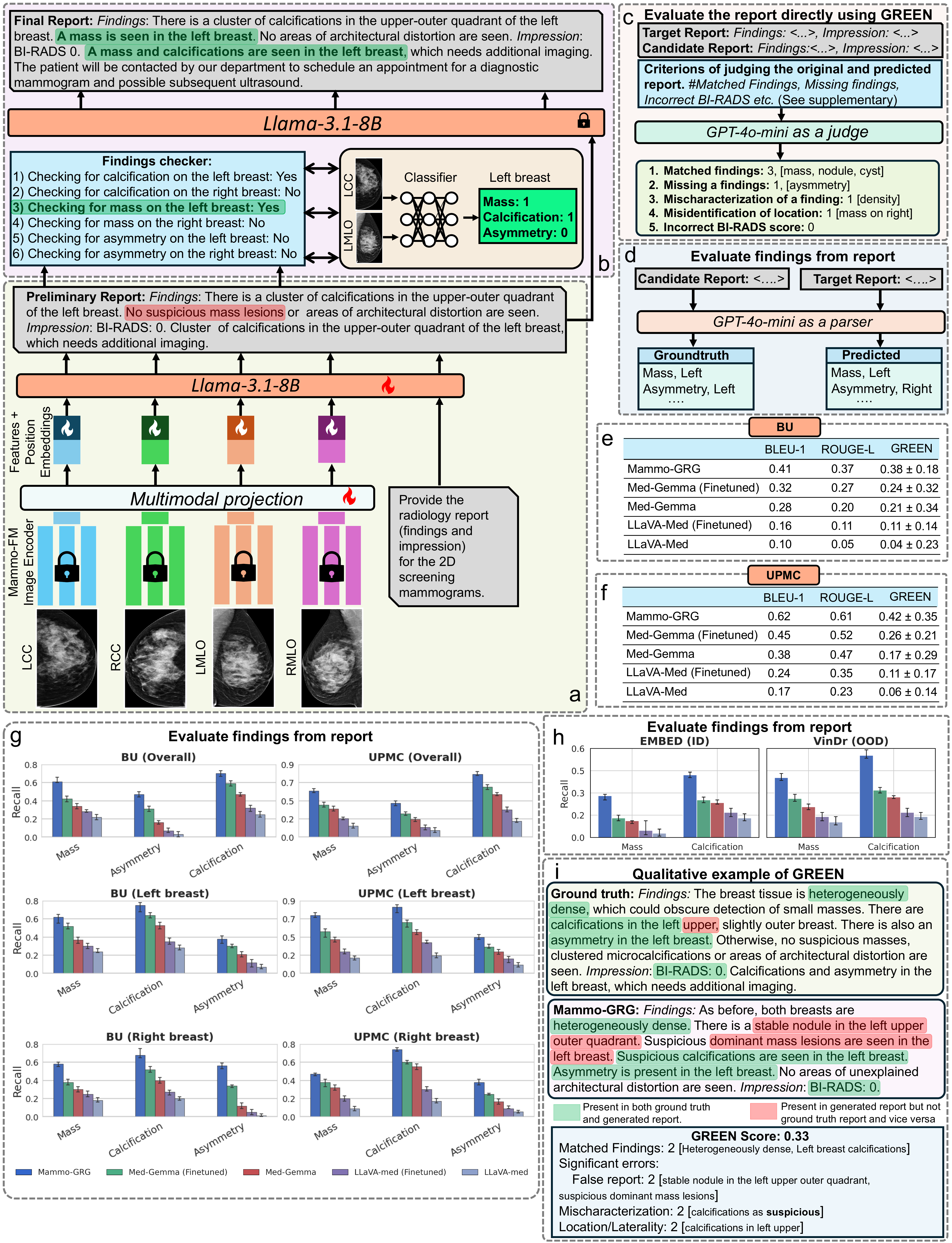}
    \vspace{-0.8em}
    \caption{\textbf{Overview of Mammo-GRG}. \textbf{a.} Schematic of the Mammo-GRG architecture. Four-view screening mammograms (LCC, LMLO, RCC, RMLO) are encoded by dedicated Mammo-FM vision encoders and projected to the latent space of a Llama-3.1-8B language model via a multimodal projector. View-specific tokens and positional embeddings preserve spatial and semantic distinctions, enabling cross-view reasoning. \textbf{b.} Clinical grounding stage where Mammo-GRG generates the final radiology report from the preliminary report. Mammo-GRG leverages Mammo-FM in zero-shot mode to predict key findings (mass, calcification, asymmetry) that constrain report generation. \textbf{c.} The GREEN evaluation framework evaluates the factual and clinical correctness of the generated report by comparing candidate and target reports using an LLM judge (GPT-4o-mini). \textbf{d.} Structured findings are extracted from both generated and reference reports using GPT-4o-mini. \textbf{e–f} Lexical (BLEU-1, ROUGE-L) and clinical (GREEN) performance on BU and UPMC datasets demonstrate consistent improvements of Mammo-GRG over fine-tuned MedGemma and LLaVA-Med baselines. \textbf{g.} Recall of key mammographic findings extracted from generated versus reference reports across datasets and laterality, showing higher clinical recall for Mammo-GRG. \textbf{h.} Pathological grounding evaluation, showing recall for mass and calcification findings extracted from generated reports compared directly to image-level ground-truth labels on the in-distribution (EMBED) and out-of-distribution (VinDr) datasets. Mammo-GRG maintains strong predictive alignment with actual imaging findings. \textbf{i.} Qualitative example illustrating the GREEN score assessment of a Mammo-GRG-generated report against ground truth, highlighting accurate recovery of findings and clinical consistency.}   
    \label{fig:report_generator}
\end{figure}

Radiology report generation is a critical task in breast imaging, yet the screening mammography domain lacks a clinically grounded, multi-view model that can generate reliable reports. In this paper, we develop  Mammo Grounded Report Generator (Mammo-GRG), the first multi-view, clinically grounded report generator for screening mammography.
Mammo-GRG integrates the pre-trained Mammo-FM (multi-institution) vision encoder, a multimodal projector, and a Llama-3.1-8B LLM~\cite{grattafiori2024llama}. Mammo-GRG training follows a two-stage method (Fig.~\ref{fig:report_generator}a and Methods): 1) a vision-language alignment stage where we train only the projector 2) an end-to-end instruction tuning stage where we finetune the projector and LLM on 359,746 question-answer (QA) pairs. Mammo-GRG overcomes three core challenges: multi-view integration, data imbalance, and clinical grounding.

First, to handle the four mammogram views (LCC, LMLO, RCC, RMLO), Mammo-GRG uses parallel Mammo-FM encoders. It projects each view representation into the LLM latent space, encapsulates it within a view-specific token (\texttt{<LCC>}, \texttt{<LMLO>}, \texttt{<RCC>}, \texttt{<RMLO>}), and adds a learnable positional embedding to preserve spatial identity. Second, to mitigate data imbalance, GPT-4o parses the reports to create structured labels for key findings (\eg mass, calcification). We assign a binary indicator if at least one finding is present and compute inverse-frequency sampling weights to ensure balanced training. Third, Mammo-GRG integrates a grounding step to ensure clinical fidelity. In this step, Mammo-FM operates in zero-shot mode to predict key mammographic findings from the input mammograms (Tab.~\ref{tab:mammo_grg_grounding_prompt}). This ensures that textual descriptions remain consistent with image-derived evidence.
Refer to Tab.~\ref{tab:prompt_mammo_instruct} and~\ref{tab:prompt_multitype} for prompts for QA generation. Refer to Tab.~\ref{tab:mammo_grg_system_prompt} and~\ref{tab:mammo_grg_grounding_prompt} for the system prompt to generate the preliminary report using LLM after end-to-end instruction tuning (stage 2) and final report after the clinical grounding stage, respectively. We utilize the reports from UPMC and BU datsets to generate QA pairs and train Mammo-GRG (Fig.~\ref{fig:app_mammo_instruct} for QA distribution).

%  GREEN decompose - write about green
% (final report after the clinical grounding stage with zero-shot mammo-FM) 
We rigorously evaluate report generation across lexical, clinical, and pathological axes. For lexical quality, we report BLEU-1 and ROUGE-L to quantify overlap between generated and ground-truth reports. As lexical metrics fail to capture clinical validity, we modify the GREEN metric~\cite{ostmeier2024green} for mammography and use GPT-4o-mini as the judge to measure clinical correctness (Fig.~\ref{fig:report_generator}c; Tab.~\ref{tab:green_metric_prompt}; Methods). To measure pathological accuracy, we perform two complementary evaluations. For report \vs report evaluation, we first parse generated and reference reports from BU and UPMC into structured findings (\eg mass, calcification, asymmetry) and compute recall (Fig.~\ref{fig:report_generator}d). Next for report \vs label evaluation, Mammo-GRG generates reports for EMBED (ID) and VinDr (OOD) datasets. Next, we parse them into structured labels and compare them with each dataset’s ground-truth findings (mass and calcification). VinDr serves as the OOD benchmark and examines robustness under domain shift. GPT-4o-mini performs the parsing in both settings.
We benchmark Mammo-GRG against MedGemma-27B and LLaVA-Med. As these baselines accept only a single image, we generate reports for each view independently and concatenate them. To isolate the impact of the vision encoder, we create two additional baselines: MedGemma (Finetuned) and LLaVA-Med (Finetuned). These variants substitute the Mammo-FM vision encoder with the encoders from MedGemma-27B and LLaVA-Med, respectively, keeping all other architectural components of Mammo-GRGconstant. In all evaluations, the `generated report' refers to the final output after the clinical grounding stage (Fig.~\ref{fig:report_generator}b).

Mammo-GRG outperforms all the baselines across all evaluation settings. Fig. \ref{fig:report_generator}e–f show consistent gains in BLEU-1, ROUGE-L, and -- most importantly -- GREEN scores over all baselines. On UPMC, the model achieves a GREEN score of $0.42 \pm 0.35$, yielding a 62\% improvement over MedGemma (Finetuned) ($0.26 \pm 0.21$) and a 6-fold increase relative to LLaVA-Med (Finetuned) ($0.06 \pm 0.14$). The BU dataset exhibits the same performance trend. Next, Mammo-GRG demonstrates superior performance in identifying key mammographic findings from the generated report (Fig.~\ref{fig:report_generator}g). On the BU dataset, Mammo-GRG achieves higher recall than the generalist fine-tuned MedGemma by 45\% for masses (0.61 \vs 0.42), 18\% for calcifications (0.70 \vs 0.59), and 52\% for asymmetry (0.47 \vs 0.31) ($P \textless 0.001$ for all). On the UPMC dataset, it achieves similar recall improvements of 26–45\% ($P<0.001$ for all) across the same findings.
Refer to Fig.~\ref{fig:app_report_accuracy} for analogous comparisons using the accuracy metric. Also, refer to Fig.~\ref{fig:app_birads_metric} for comparing BI-RADs scores.
We evaluate generated reports against ground-truth finding labels (Fig.~\ref{fig:report_generator}h).
On the OOD VinDr dataset, Mammo-GRG demonstrates strong robustness a 103\% and 95\% higher recall for suspicious calcifications (0.59 \vs 0.29) and masses (0.43 \vs 0.22), respectively ($P = 0.004$ for both), compared to the base MedGemma model.
A similar trend follows on the ID EMBED dataset where Mammo-GRG outperforms fine-tuned MedGemma by 68-114\% for calcification and mass, respectively. Fig.~\ref{fig:app_mammo_grg_ft} details the performance gain achieved by the clinical grounding step, comparing the final grounded report to the preliminary report (post-instruction tuning). Together, these results establish Mammo-GRG as the first clinically grounded, multi-view mammography report generator with robust performance across all evaluation axes.

\label{subsec:report_gen}

\section{Discussion}

Mammo-FM (multi-institution) is trained on a large, geographically and demographically diverse corpus of 140,677 patients from UPMC, BU, Mayo Clinic, and EMBED, enabling strong cross-domain generalization. To preserve institutional privacy, the training protocol adopts a simple federated learning setup, alternating training epochs between a central server (with UPMC, BU and EMBED data) and the Mayo Clinic site. This protocol avoids the synchronization cost of more complex algorithms \eg FedAvg, which remain viable for future exploration. Our pretraining data consists of screening mammograms (BI-RADS 0, 1, and 2 only) and associated reports, driving the model to learn the semantic precursors to malignancy -- \eg masses and calcifications -- directly from the text. Our Mammo-FM architecture addresses mammography’s unique visual and linguistic constraints. Its EfficientNet-B5 backbone supports higher resolution $1520 \times 912$ mammograms, retaining the fine-grained essential diagnostic details\eg calcifications. The ModernBERT text encoder accommodates the long-form radiology reports (mean 74 tokens, max 505) without truncation, preserving local and contextual semantics. We further initialize ModernBERT by fine-tuning on $\sim$200k chest CT radiology reports, substantially enhancing multimodal alignment and downstream clinical performance.

% Data Efficiency and Generalizability:
Mammo-FM demonstrates strong generalization across both ID (EMBED) and OOD (RSNA and VinDr) datasets. The multi-institutional variant consistently exceeds the UPMC-only model by 13–15\% AUROC on average, confirming the importance of data diversity and scale for robust representation learning. Notably, performance remains high on VinDr dataset -- a Vietnamese cohort -- demonstrating robustness to geographic, demographic, and scanner shifts outside the pretraining domain.
Among baselines, MedSigLIP’s broad medical pretraining performs better than DINOv3’s natural-image features, while CXR-CLIP's narrow domain specialization performs the worst. Our evaluation focuses on mass, calcification, and architectural distortion, as other findings have insufficient prevalence (\textless1\%, Tab.~\ref{tab:dataset_eval_characteristics}).  
% Across individual findings, calcification classification achieves the highest AUROC (\textless 0.95), as calcifications present with distinct high-contrast morphology, consistent terminology in reports, and limited inter-case variability -- factors that facilitate robust visual–textual alignment. 
Full fine-tuning (only for 5 epochs) consistently surpasses linear probing, since end-to-end optimization enables deeper adaptation of learned representations to task-specific distributions. In both settings, Mammo-FM (multi-institution) trained on just 25\% of the data outperforms MedSigLIP trained on 80\%, achieving on average a 10–15\% higher AUROC, consistently across ID and OOD benchmarks. Note that Mammo-FM image encoder is 10-fold smaller than MedSigLIP image encoder($\sim$30M \vs400M parameters). 
Such high performance emerges from (1) pretraining on the largest real-world mammography–report corpus to date, rather than general medical or natural image data, and (2) processing high-resolution mammograms ($1{,}520 \times 912$), preserving fine-scale diagnostic patterns such as microcalcifications that encoders with low image size (MedSigLIP: $448 \times 448$; DINOv3: $512 \times 512$; CXR-CLIP: $224 \times 224$) fail to capture.

% prognosis
Unlike MIRAI and AsymMIRAI, Mammo-FM-integrated breast cancer risk models enable both image and text-based interpretability due to Mammo-FM's vision-language alignment.
For post hoc concept-level interpretation, we employ a Matryoshka Sparse Autoencoder (SAE) to retain the expressivity of the original predictor (Fig.~\ref{fig:prognosis}h). This framework, for the first time, grounds visual activations in corresponding radiology report sentences, enabled by Mammo-FM’s shared image–text embedding space. 
We train our Mammo-FM-based risk predictors using knowledge distillation (KD) from MIRAI. We select MIRAI as the teacher as its training on large-scale cancer datasets enables it to encode robust malignancy patterns. In contrast, Mammo-FM is pretrained on screening mammograms without explicit cancer supervision. This distillation transfers MIRAI’s prognostic malignancy signal into Mammo-FM-based risk predictors while retaining the latter’s image-text alignment. Between the two variants, MIRAI w/ Mammo-FM outperforms AsymMIRAI w/ Mammo-FM due to its transformer-based aggregation of multi-view features and concurrent training across all risk horizons.

% report generation
Evaluating medical report generation remains a major challenge, as conventional linguistic metrics such as BLEU and ROUGE fail to capture clinical accuracy. Therefore, for the first time, we adapt the GREEN framework, using an LLM judge to quantify mammography-specific clinical factuality and serve as a stronger benchmark for radiology report assessment. 
Mammo-GRG, the first multi-view report generator for mammography, outperforms all generalist baselines by achieving $\textgreater$55\%  higher GREEN scores than the fine-tuned generalist model MedGemma. Mammo-GRG consistently performs better on UPMC than BU, due to richer linguistic structure and longer contextual descriptions in UPMC reports (mean sentences 47.5 in UPMC  \vs 32.1 in BU, refer to Fig.~\ref{fig:app_report_stats}). Overall Mammo-GRG's high performance is driven by explicit architectural and data-level design. First, Mammo-GRG is instruction-tuned on 359,746 QA pairs derived from real-world mammography reports across UPMC and BU, providing task-specific supervision that captures breast-imaging semantics instead of generic medical language. Second, unlike the baselines, our Mammo-FM vision encoder processes all four high-resolution ($1520 \times 912$) views simultaneously, using view-specific tokens and positional embeddings, mirroring the clinical reasoning process. Finally, Mammo-GRG integrates Mammo-FM’s high-fidelity vision–language representations for factual verification (refer to Fig.~\ref{fig:app_Mammo_GRG_grounding} for the effect of Mammo-FM's zero-shot grounding on Mammo-GRG), ensuring generated findings remain anatomically and clinically consistent.

% limnitation:
Despite these advances, our study has several limitations. First, our multi-institutional pretraining dataset may inherit demographic or scanner biases. Second, the risk predictor may inherit latent biases from the teacher model -- MIRAI. Third, zero-shot classification remains sensitive to prompt variation. Fourth, the performance of Mammo-GRG depends on the capabilities and alignment of its underlying large language model (LLM). The LLM occasionally hallucinates findings or misinterprets stable lesions that depend on prior comparisons that are absent from the current input. Finally, our framework is limited to 2D mammograms; extending it to 3D modalities (DBT or MRI) could better capture volumetric and temporal disease patterns.

In summary, this paper establishes that domain-specific foundation models achieve performance, interpretability, and clinical value beyond the reach of generalist models. Tailored image-text models like Mammo-FM and Mammo-GRG represent a clear path toward transparent, trustworthy, and scalable AI in medicine.

\label{subsec:discussion}

\clearpage

\bibliographystyle{plainnat}
\bibliography{refs}

\clearpage
\appendix

\section*{Supplementary Materials}
\subsection*{Dataset}
We compile a large-scale, multi-institutional pre-training corpus from four distinct sources: three private institutional datasets from the Mayo Clinic, the University of Pittsburgh Medical Center (UPMC), and Boston Medical Center, the primary teaching affiliate of the Boston University (BU) School of Medicine,  and the public, and the public Emory Breast Imaging Dataset (EMBED)~\cite{jeong2023emory}. We filter all datasets to include only two-dimensional (2D) screening mammograms with a Breast Imaging Reporting and Data System (BI-RADS) assessment of 0, 1, or 2. Also, we select the mammograms with only CC and MLO views, excluding all other views. All dataset partitions are performed at the patient level to ensure no overlap of patient data across training, validation, or test splits.

\textbf{Mayo Clinic Dataset.} This private institutional dataset consists of screening mammograms and their corresponding radiology reports (Tab.~\ref{tab:main_dataset_char}). From a total of 96,567 patients and 465,078 images, we create a training set of 77,253 patients (372,248 images) and a held-out validation set of 9,657 patients (46,298 images).

\begin{figure*}[h]
\begin{center}
\includegraphics[width=0.88\textwidth]{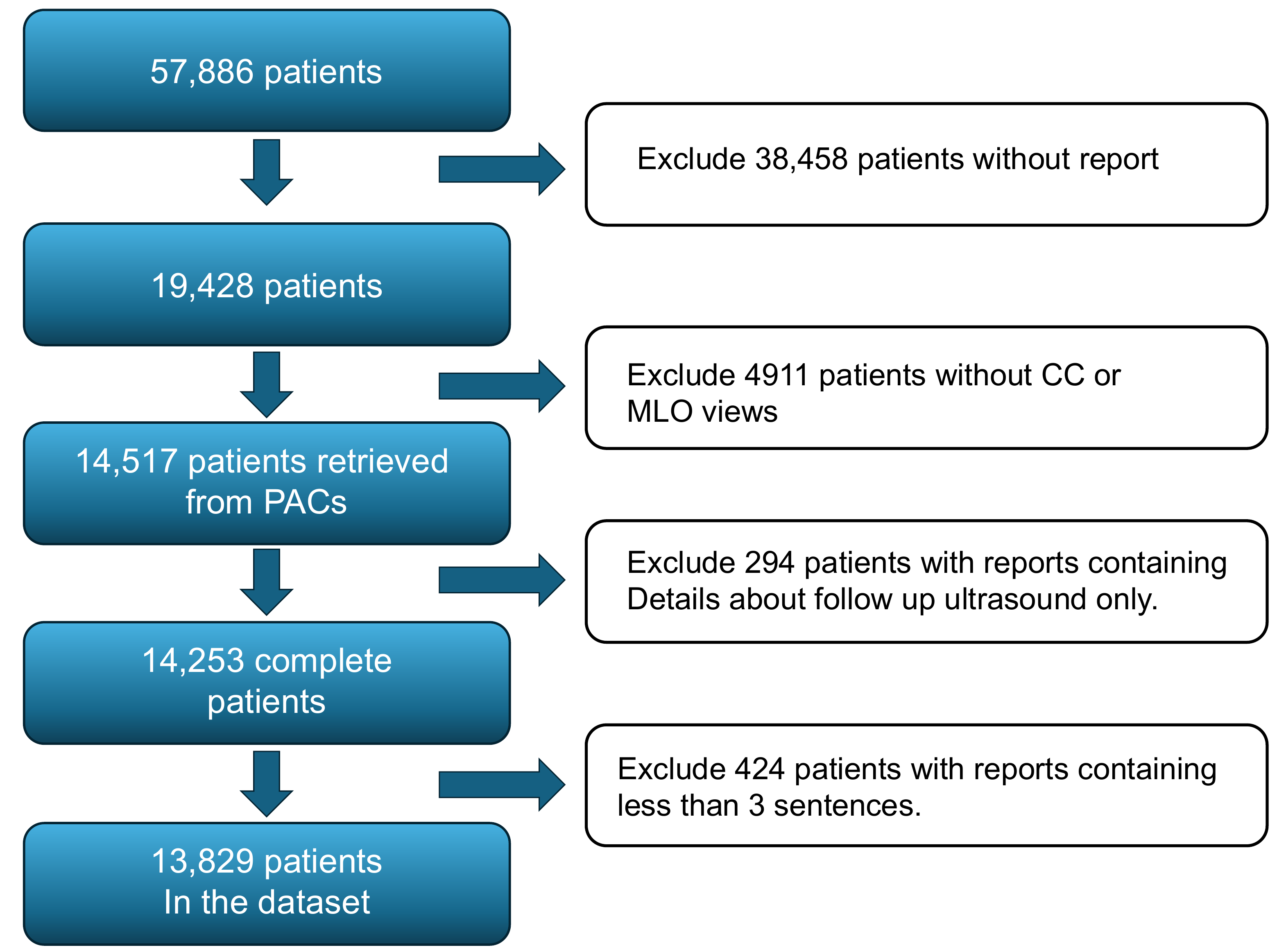}
\caption{\textbf{Selection pipeline of the UPMC dataset used for pretraining Mammo-FM}}
\label{fig:app_datapreprocessing_UPMC}
\end{center}
\end{figure*}

\textbf{University of Pittsburgh Medical Center (UPMC) Dataset.} This private institutional dataset contains 46,433 screening images and 13,829 associated reports from 13,829 patients (Fig.~\ref{fig:app_datapreprocessing_UPMC}, Tab.~\ref{tab:main_dataset_char}). We partition the dataset into a training set (6,921 patients; 26,543 images), a validation set (3,461 patients; 10,123 images), and a test set (3,447 patients; 9,857 images). We utilize this test set to evaluate our report generation module.

\begin{figure*}[h]
\begin{center}
\includegraphics[width=0.88\textwidth]{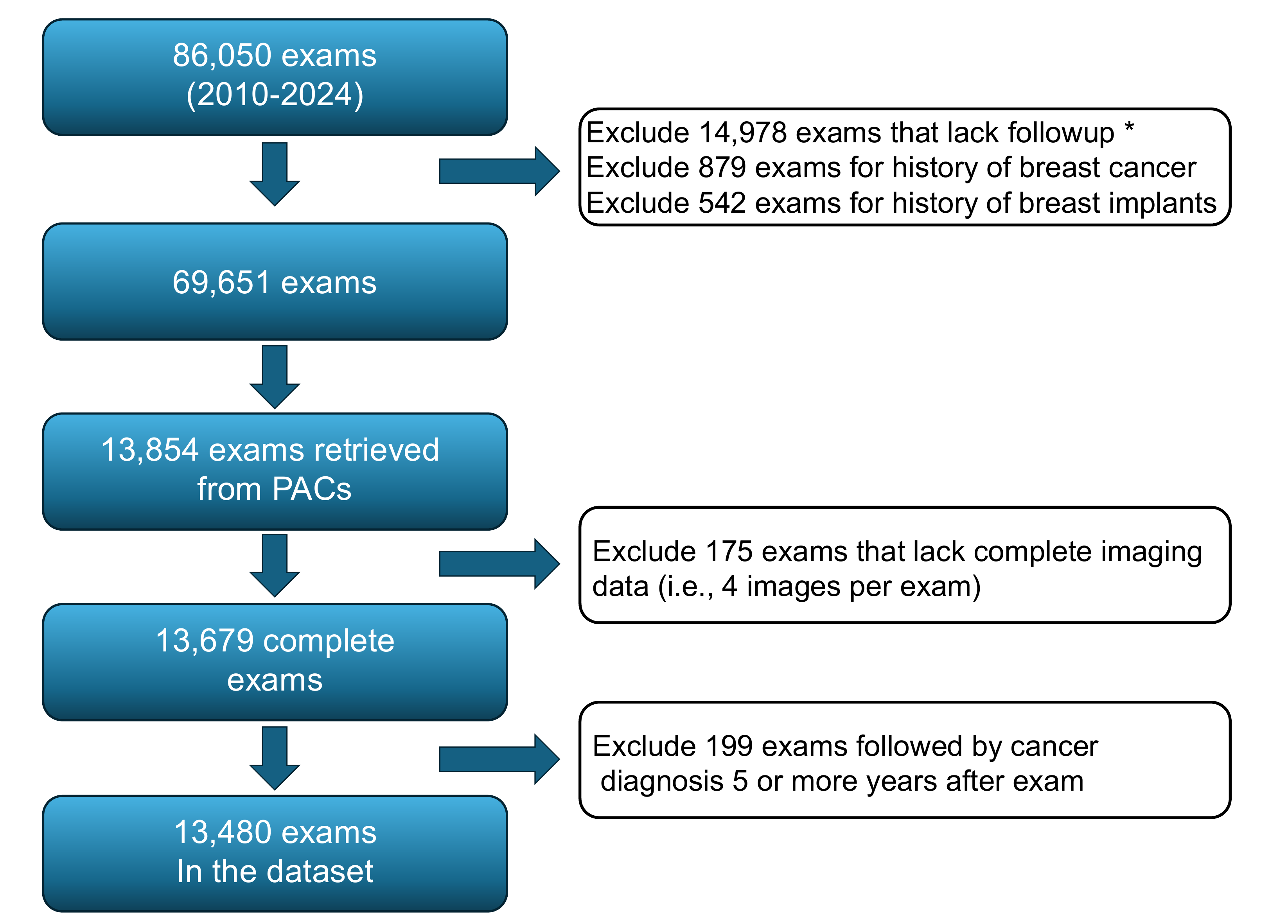}
\caption{\textbf{Selection pipeline of the BU dataset used for pretraining Mammo-FM}}
\label{fig:app_datapreprocessing_BU}
\end{center}
\end{figure*}

\textbf{Boston University (BU) Dataset.} This private institutional dataset from  Boston Medical Center includes 54,776 screening mammograms from 9,900 patients. It consists of 13,480 exams with complete data that includes associated radiology reports and followup data on clinical outcomes (Fig.~\ref{fig:app_datapreprocessing_BU}, Tab.~\ref{tab:bu_dataset_characteristics}) We split the data into a training set (4,950 patients), a validation set (2,475 patients), and a test set (2,475 patients). Similar to the UPMC dataset, we utilize this test set to evaluate our report generation module.

\begin{figure*}[h]
\begin{center}
\includegraphics[width=0.88\textwidth]{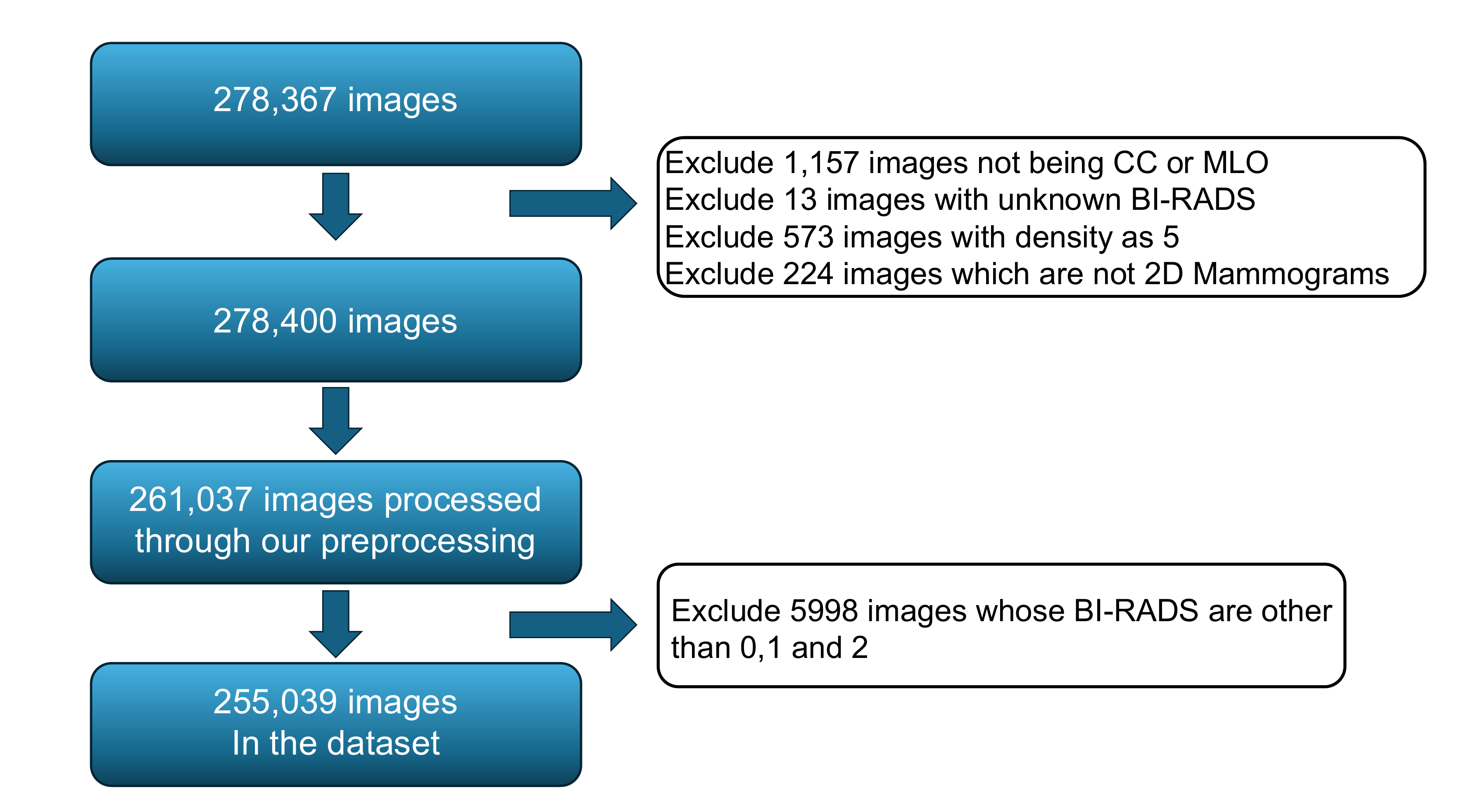}
\caption{\textbf{Selection pipeline of the EMBED dataset used for pretraining Mammo-FM}}
\label{fig:app_datapreprocessing_EMBED}
\end{center}
\end{figure*}

\textbf{Emory Breast Imaging Dataset (EMBED) Dataset.} EMBED is a large mammography dataset from Emory, Georgia in the US. This paper utilizes only the publicly available cohort of EMBED.
Unlike our other pre-training datasets, it contains structured tabular annotations rather than free-text radiology reports (Fig.~\ref{fig:app_datapreprocessing_EMBED}, Tab.~\ref{tab:main_dataset_char}). Instead, it provides structured mammographic attributes for findings \eg \textit{mass}, \textit{calcifications} \etc. We utilize a subset of this public resource, filtering to include only 2D screening mammograms with a BI-RADS assessment of 0, 1, or 2. This results in a cohort of 20,381 patients and 255,039 images, which we partition into a training set (10,190 patients; 126,884 images) and a validation set (5,085 patients; 64,029 images). We allocate the remainder as a test set for evaluating the representations of Mammo-FM.

\subsection*{Additional results}
\begin{figure*}[h]
\begin{center}
\includegraphics[width=0.88\textwidth]{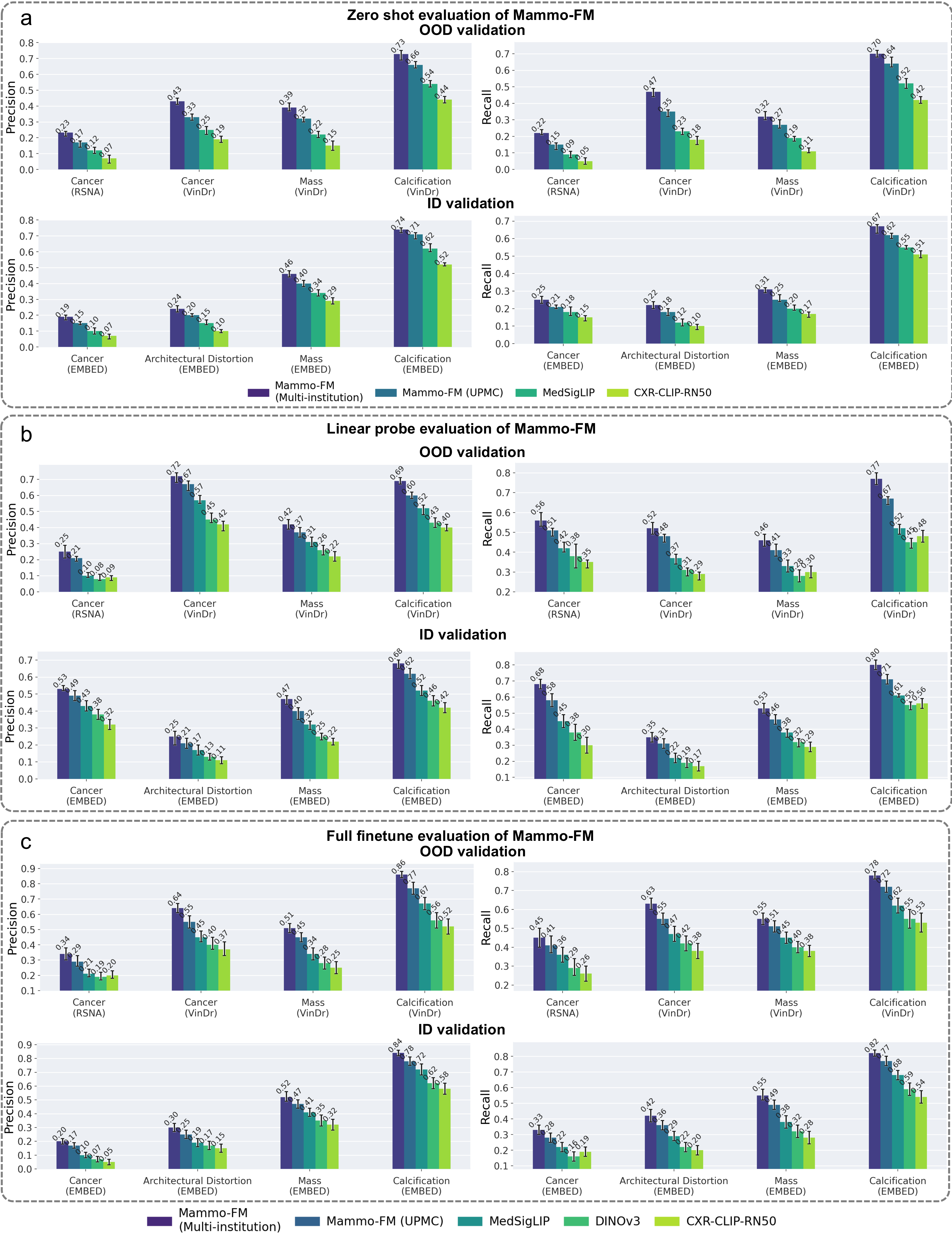}
\caption{\textbf{Evaluation (Precision and Recall) of diagnostic performance across training regimes and validation settings.} \textbf{a.} Zero-shot evaluation of Mammo-FM and baseline models across out-of-distribution (OOD) and in-distribution (ID) validation sets.
\textbf{b.} Linear probe evaluation of the same models, where only the classification layer was trained on labeled data while image–text encoders remained frozen.
\textbf{c.} Full fine-tuning evaluation of Mammo-FM and comparator models.
Each bar represents mean precision and recall across diagnostic categories (Cancer, Architectural distortion, Mass, and Calcification), with error bars denoting standard deviation across validation folds.
Across all regimes, Mammo-FM (Multi-institution) consistently achieves the highest diagnostic precision and recall, demonstrating strong generalization across datasets compared with domain-specific and vision-only baselines (MedSigLIP, DINOv3, and CXR-CLIP-RN50)}
\label{fig:app_diagnosis_precision_recall}
\end{center}
\end{figure*}

\begin{figure*}[h]
\begin{center}
\includegraphics[width=0.88\textwidth]{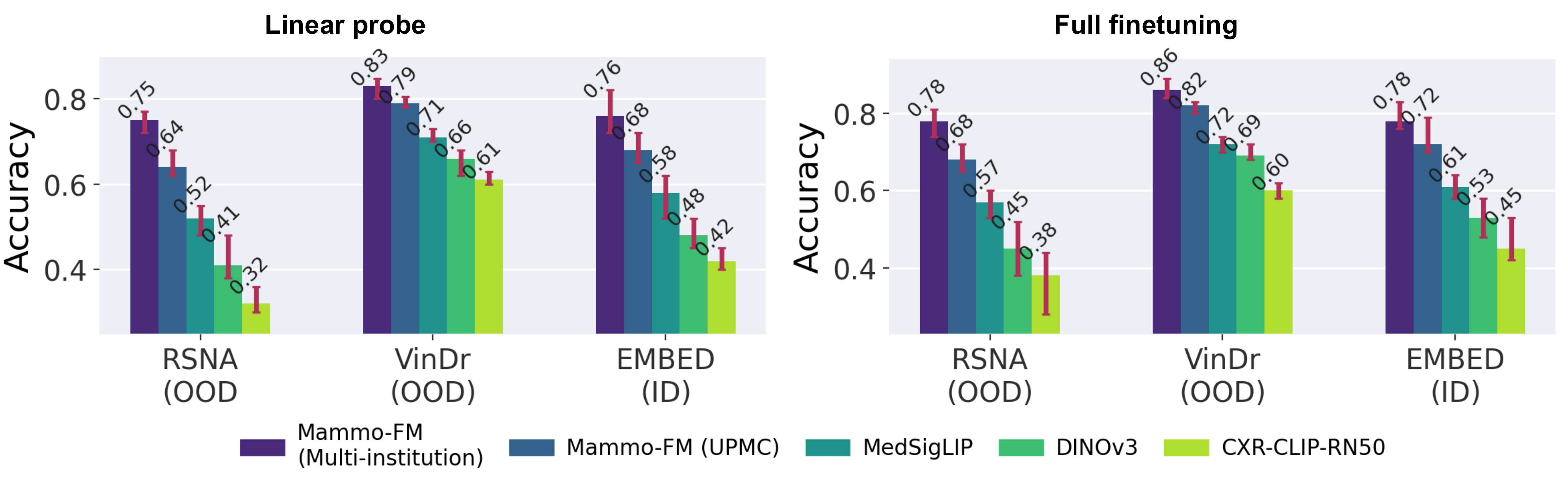}
\caption{\textbf{Accuracies of breast density classification across pre-trained all the image encoders.}
Mammo-FM (multi-institution) achieves the highest accuracy across both in-distribution (EMBED) and out-of-distribution (RSNA, VinDr) datasets under linear probe and full fine-tuning settings. Its consistent gains over Mammo-FM (UPMC), MedSigLIP, DINOv3, and CXR-CLIP-RN50 highlight the benefits of large-scale, domain-specific pretraining and multi-institutional data diversity for robust mammographic representation learning.}
\label{fig:app_density}
\end{center}
\end{figure*}

\begin{figure*}[h]
\begin{center}
\includegraphics[width=0.88\textwidth]{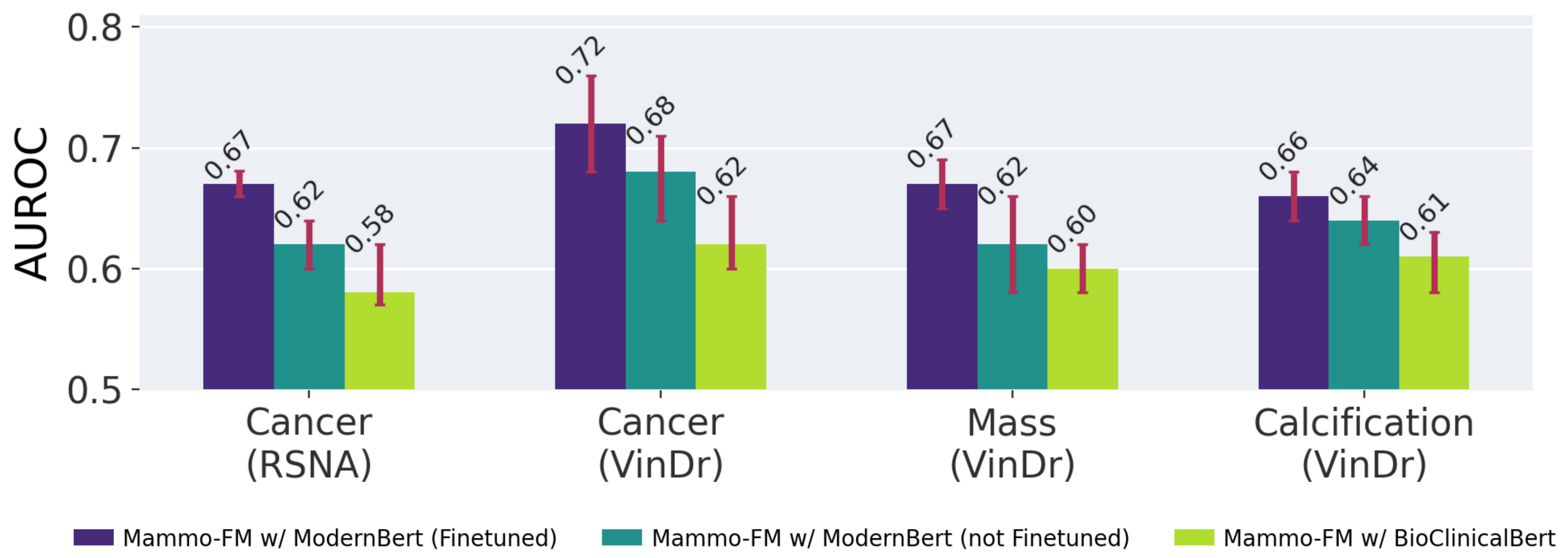}
\caption{\textbf{Zero-shot AUROC performance across out-of-distribution (OOD) datasets using different text encoders of Mammo-FM pretrained on the UPMC dataset.}
The plot compares multiple text encoder variants under zero-shot settings, highlighting the impact of the language backbone. Finetuned and non-finetuned ModernBERT variants indicate whether the text encoder is finetuned on 200,000 UPMC radiology reports, which serve as the initialization for the Mammo-FM model. Finetuned variants show stronger visual–textual alignment and higher cross-domain robustness compared to non-finetuned counterparts.}
\label{fig:app_text_encoder}
\end{center}
\end{figure*}

\begin{figure*}[h]
\begin{center}
\includegraphics[width=0.88\textwidth]{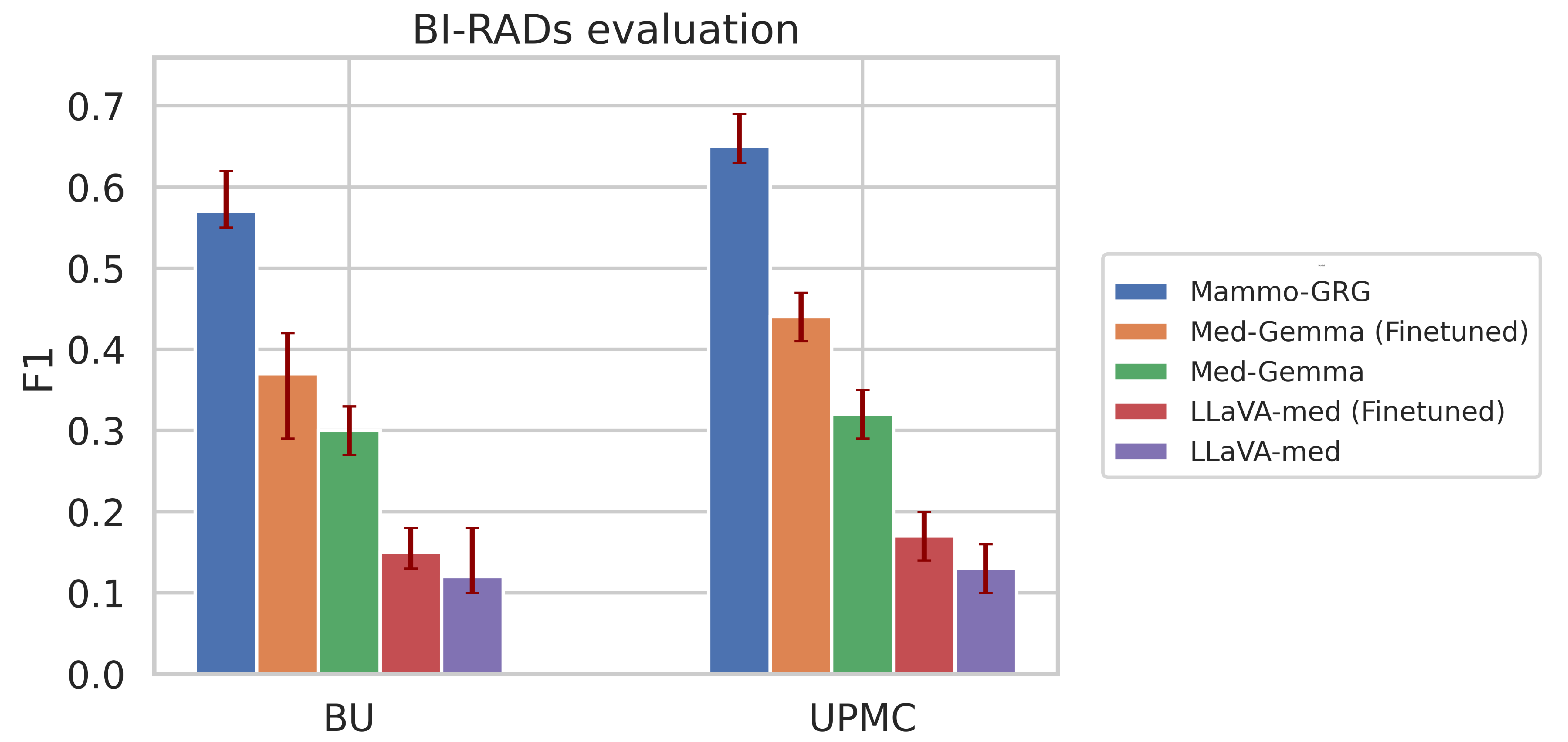}
\caption{\textbf{Accuracies of BI-RADS classification performance across report generation models: Mammo-GRG, Med-Gemma (fine-tuned), Med-Gemma, LLaVA-med (fine-tuned), and LLaVA-med.} Across both the datasets -- UPMC and BU, Mammo-GRG demonstrates superior performance compared with all generalist baselines, indicating more clinically reliable BI-RADS classification from generated reports. For this evaluation, we group BI-RADS 0 as ``suspicious'' and merge BI-RADS 1–2 as ``negative/benign,'' computing the F1-score between predicted and ground-truth categories. }
\label{fig:app_birads_metric}
\end{center}
\end{figure*}

\begin{figure*}[h]
\begin{center}
\includegraphics[width=0.88\textwidth]{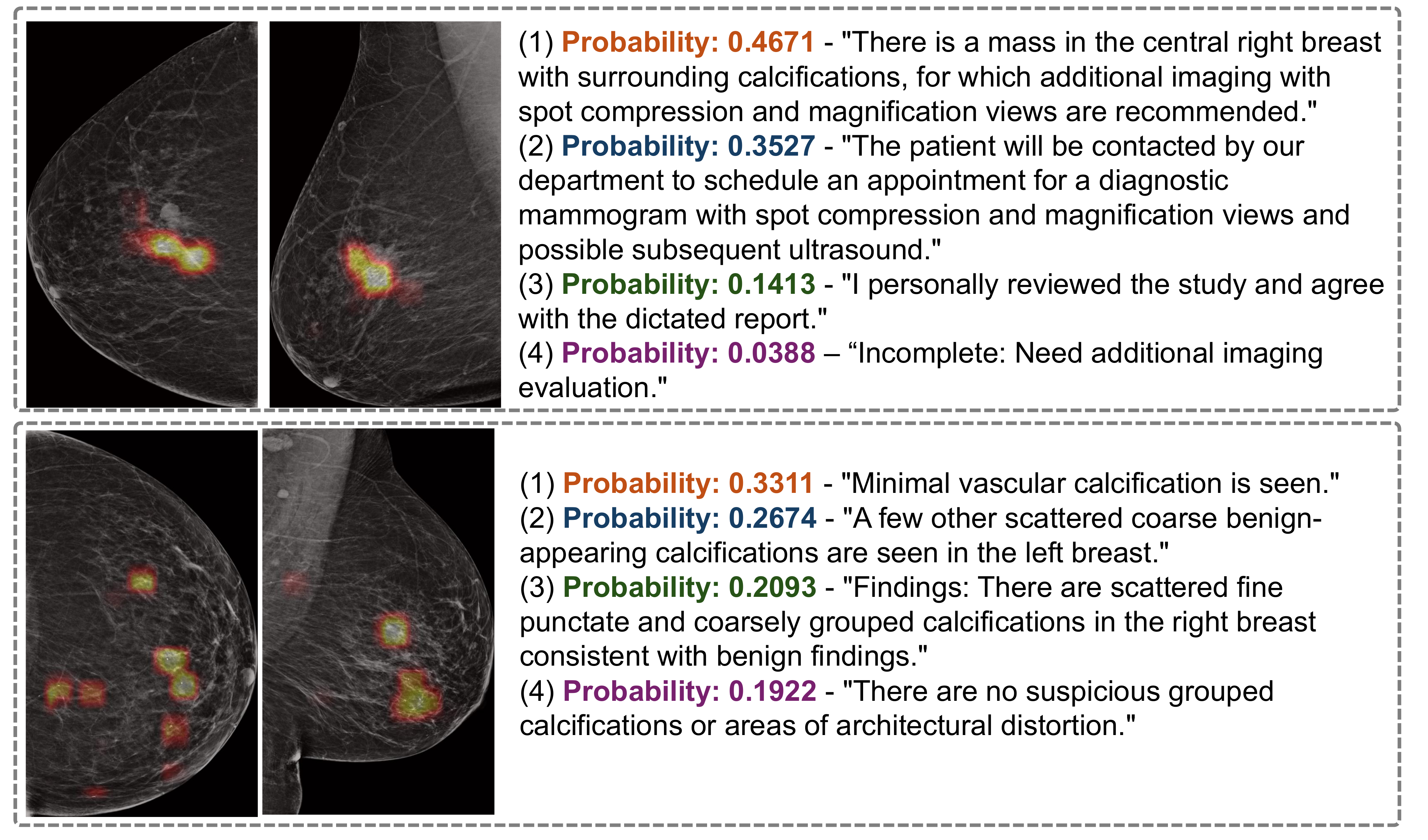}
\caption{\textbf{Qualitative examples of text-grounded interpretability of the MIRAI w/ Mammo-FM risk predictor on BU dataset samples for 1-year cancer risk prediction.} Each heatmap (left) localizes high-risk activations driving the model’s prediction. We rank the report sentences (right) by probabilities derived from the causal ablation distribution, computed as the normalized drop in cosine similarity between the original and neuron-ablated image–text embeddings. High-probability sentences capture the dominant clinical rationale behind the predicted risk. This alignment emerges as the risk predictor uses the Mammo-FM image encoder, whose shared image–text embedding space enables direct linguistic grounding of visual evidence—unlike prior risk models lacking such multimodal alignment.}
\label{fig:app_text_interpretability}
\end{center}
\end{figure*}

\begin{figure*}[h]
\begin{center}
\includegraphics[width=0.88\textwidth]{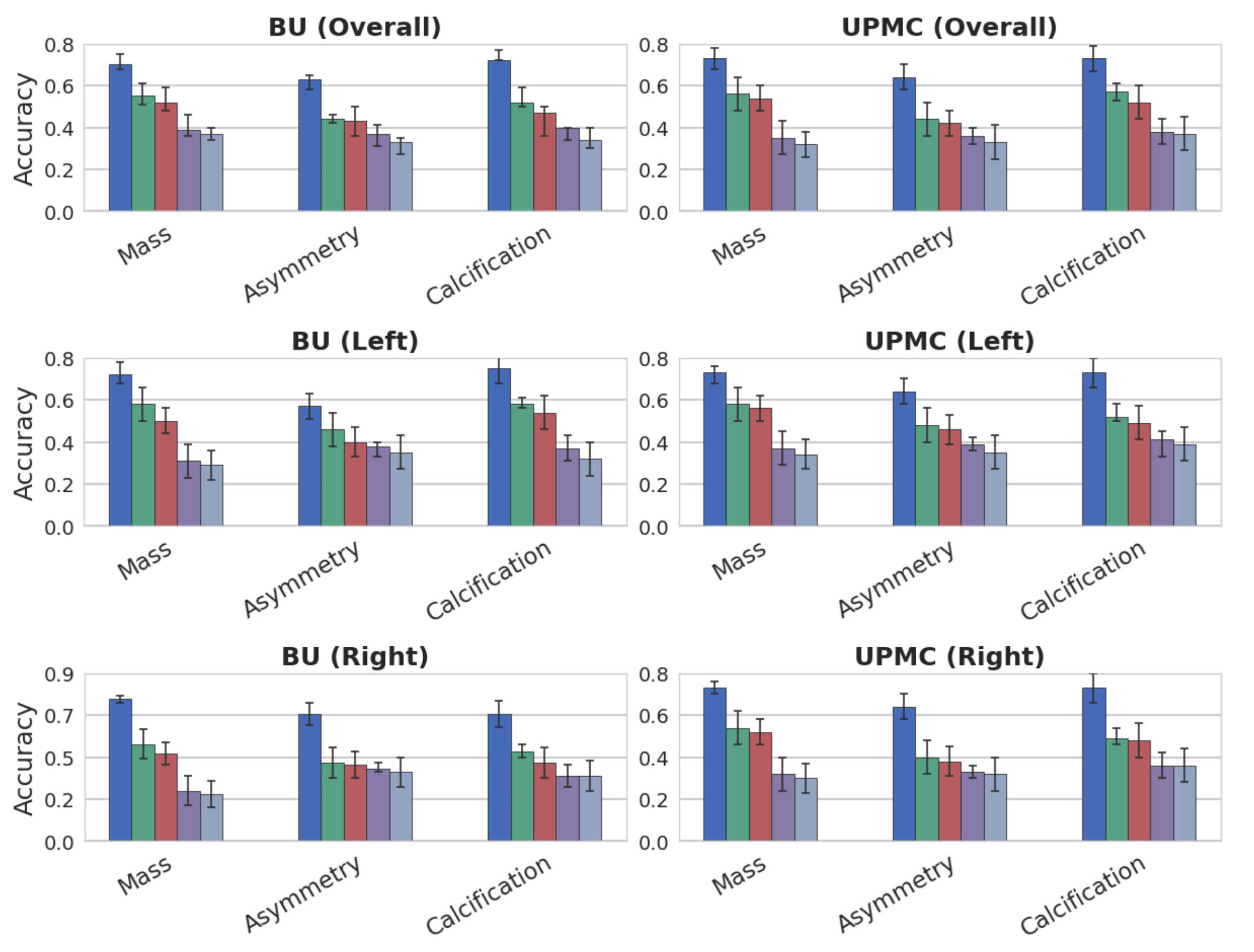}
\caption{\textbf{Accuracies of key mammographic findings (\eg mass, calcification, and asymmetry) extracted from generated versus reference reports across datasets and laterality.} Across both datasets (UPMC and BU) and for both left and right breasts, Mammo-GRG consistently outperforms all generalist baselines, demonstrating superior accuracy and more clinically reliable BI-RADS classification from generated reports.}
\label{fig:app_report_accuracy}
\end{center}
\end{figure*}

\begin{figure*}[h]
\begin{center}
\includegraphics[width=0.88\textwidth]{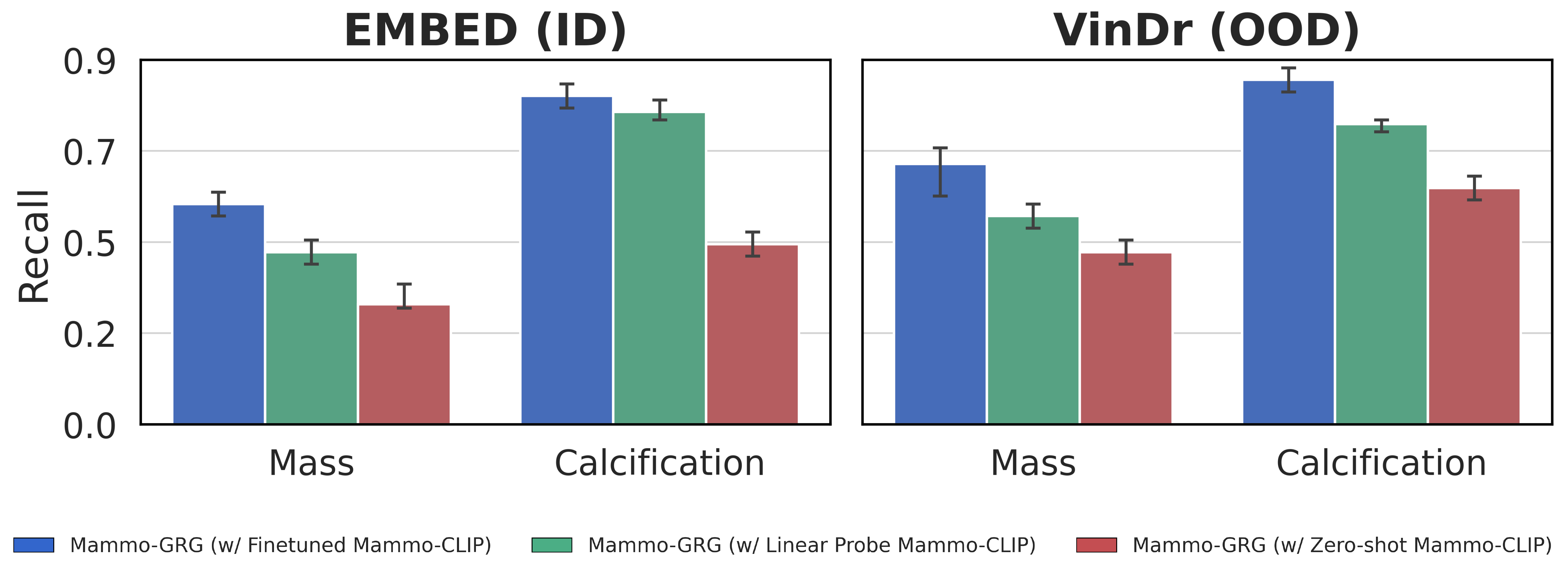}
\caption{\textbf{Comparison of Mammo-GRG fact-checking performance using different Mammo-FM configurations.} Evaluation of Mammo-GRG report verification using fully fine-tuned (FT), linear probe (LP), and zero-shot (ZS) variants of Mammo-FM across in-distribution (EMBED) and out-of-distribution (VinDr) datasets. Recall is reported for mass and calcification detection. The fully fine-tuned Mammo-FM achieves the highest recall but requires extensive labeled data and full model optimization, making it computationally expensive. The linear probe approach, while less costly, still depends on labeled supervision. The zero-shot Mammo-FM offers the most efficient and label-free alternative, demonstrating strong generalization and competitive performance across datasets.}
\label{fig:app_mammo_grg_ft}
\end{center}
\end{figure*}

\begin{figure*}[h]
\begin{center}
\includegraphics[width=0.88\textwidth]{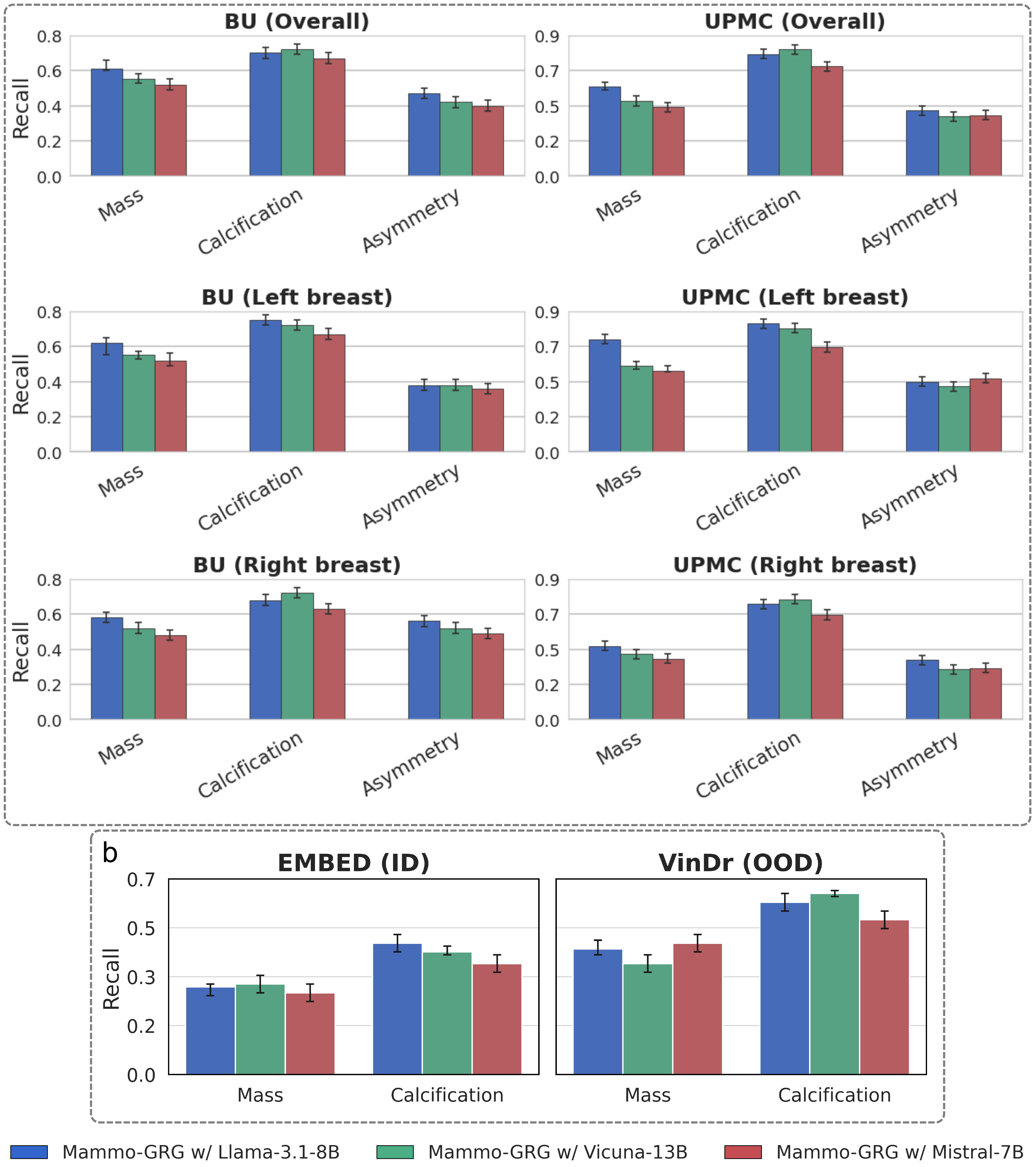}
\caption{\textbf{Ablation study of Mammo-GRG using different large language models for report generation.} \textbf{a.} Recall for key diagnostic findings (BI-RADS category, mass, calcification, and asymmetry) extracted from reports generated by Mammo-GRG models based on different LLM backbones—Llama-3.1-8B, Vicuna-13B, and Mistral-7B—is shown for the BU and UPMC datasets, separately for overall, left-breast, and right-breast evaluations. \textbf{b.} Generalization performance on in-distribution (EMBED) and out-of-distribution (VinDr) datasets, where reports were generated using each Mammo-GRG variant and parsed with GPT-4o-mini to extract structured labels for mass and calcification findings. Recall was computed relative to ground-truth labels. Across datasets, Mammo-GRG initialized with Llama-3.1-8B consistently demonstrates higher factual recall, suggesting that larger instruction-tuned LLMs improve clinical finding coverage in generated radiology reports.}
\label{fig:app_ablation_mammo_GRG_LLM_quantitative}
\end{center}
\end{figure*}

Fig.~\ref{fig:app_diagnosis_precision_recall} illustrates evaluation (Precision and Recall) of diagnostic performance across training regimes and validation settings. Fig.~\ref{fig:app_density} compares the performance of various image encoders for breast density classification across linear probe and full fine-tuning settings. Fig~\ref{fig:app_text_encoder} compares the effect of various text encoders (Bio Clinical Bert \vs Modern Bert with and without finetuning on 200k radiology report from UPMC) on the zero-shot classification performance of Mammo-FM.
Fig~\ref{fig:app_text_interpretability} shows more qualitative examples of text-grounded interpretability of the MIRAI w/ Mammo-FM risk predictor on samples of BU dataset.
Fig~\ref{fig:app_birads_metric} reports accuracies of BI-RADS classification performance across report generation models: Mammo-GRG, Med-Gemma (fine-tuned), Med-Gemma, LLaVA-med (fine-tuned), and LLaVA-med. Fig~\ref{fig:app_report_accuracy} shows the accuracy of key mammographic findings (\eg mass, calcification, and asymmetry) extracted from generated versus reference reports across datasets and laterality. Fig.~\ref{fig:app_mammo_grg_ft} demonstrates the comparison of Mammo-GRG fact-checking performance using different Mammo-FM configurations (full fine-tune \vs linear probe \vs zero-shot). Fig.~\ref{fig:app_ablation_mammo_GRG_LLM_quantitative} reports the ablation study of Mammo-GRG using different large language models for report generation. Tab.~\ref{tab:app_ablation_mammo_GRG_llms} shows the ablation study comparing the effect of different large language models on report generation performance.

\begin{figure*}[h]
\begin{center}
\includegraphics[width=0.88\textwidth]{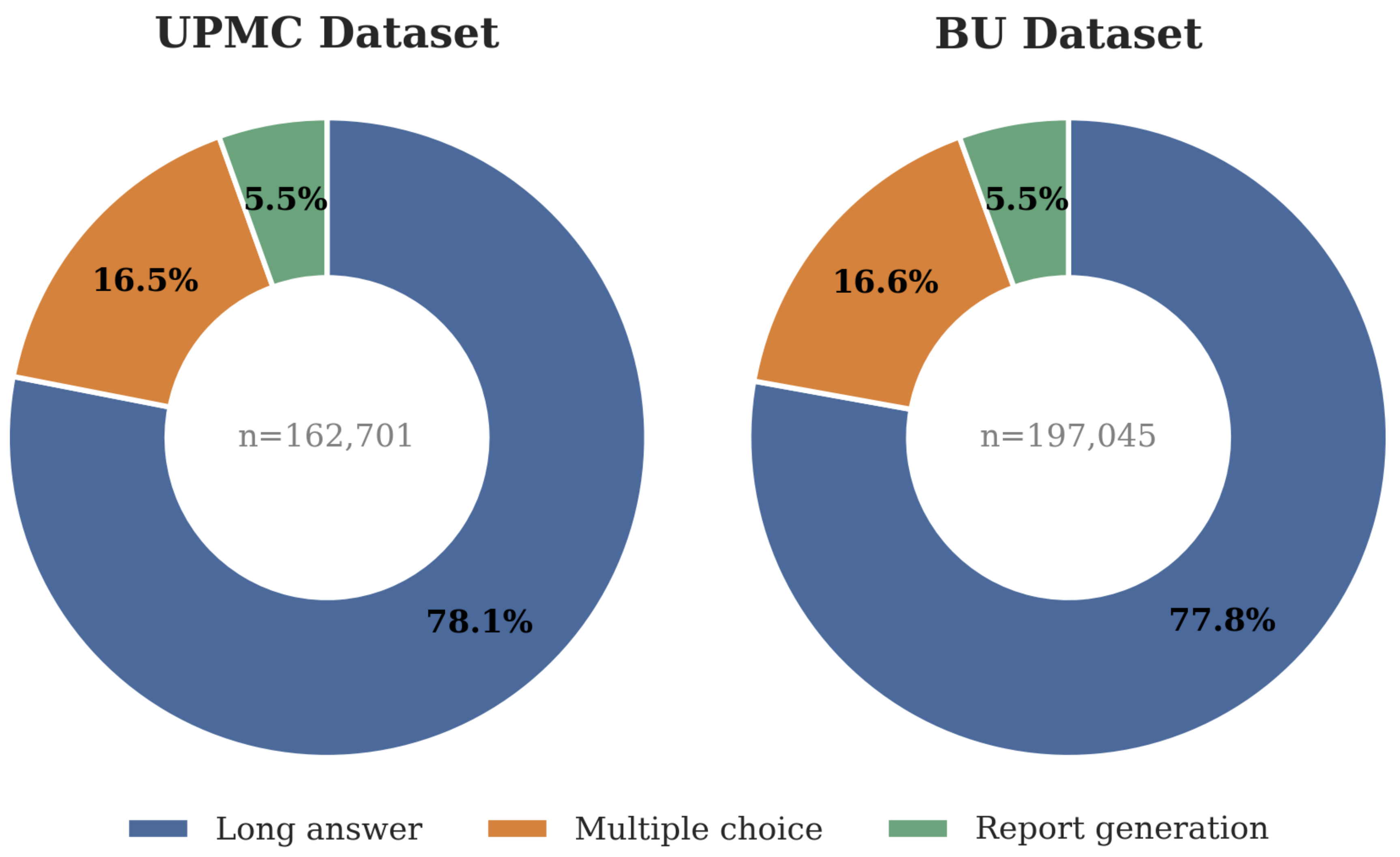}
\caption{\textbf{Distribution of question-answer (QA) pairs within the Mammo-Instruct dataset -- used to train Mammo-GRG -- across the UPMC and BU subsets.}}
\label{fig:app_mammo_instruct}
\end{center}
\end{figure*}

\begin{figure*}[h]
\begin{center}
\includegraphics[width=0.88\textwidth]{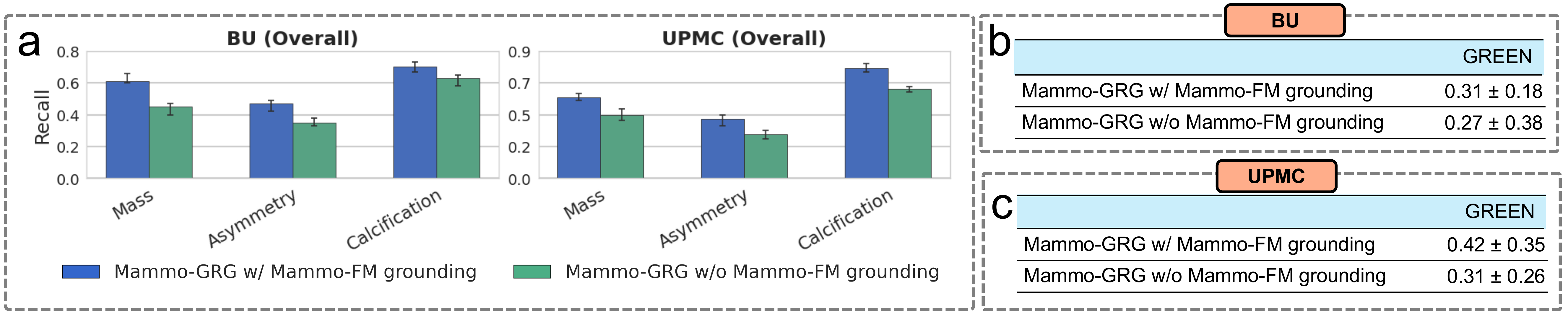}
\caption{\textbf{Effect of Mammo-FM–based zero-shot grounding on Mammo-GRG.} \textbf{a.} Recall across key clinical findings (mass, asymmetry, calcification) for BU and UPMC datasets. Mammo-GRG with Mammo-FM-zero-shot grounding achieves consistently higher recall across all categories, demonstrating improved factual accuracy and clinical relevance. \textbf{b-c.} GREEN factuality scores on BU and UPMC datasets. Mammo-GRG with Mammo-FM grounding achieves substantially higher factual alignment compared to the ungrounded version}
\label{fig:app_Mammo_GRG_grounding}
\end{center}
\end{figure*}

\begin{table}[h]
\centering
\renewcommand{\arraystretch}{1.2}
\setlength{\tabcolsep}{8pt}
% --------- BU TABLE -----------
\begin{tabular}{|>{\centering\arraybackslash}m{5cm}|c|c|c|}
\hline
\multicolumn{4}{|c|}{\cellcolor{sectionorange}\textbf{BU}} \\ \hline
\rowcolor{headerblue} 
\textbf{} & \textbf{BLEU-1} & \textbf{ROUGE-L} & \textbf{GREEN} \\ \hline
Mammo-GRG w/ Llama-3.1-8B & 0.41 & 0.37 & 0.38 $\pm$ 0.18 \\ \hline
Mammo-GRG w/ Vicuna-13B & 0.42 & 0.41 & 0.34 $\pm$ 0.24 \\ \hline
Mammo-GRG w/ Mistral-7B & 0.38 & 0.35 & 0.31 $\pm$ 0.16 \\ \hline
\end{tabular}

\vspace{1em}

% --------- UPMC TABLE -----------
\begin{tabular}{|>{\centering\arraybackslash}m{5cm}|c|c|c|}
\hline
\multicolumn{4}{|c|}{\cellcolor{sectionorange}\textbf{UPMC}} \\ \hline
\rowcolor{headerblue} 
\textbf{} & \textbf{BLEU-1} & \textbf{ROUGE-L} & \textbf{GREEN} \\ \hline
Mammo-GRG (Llama-3.1-8B) & 0.62 & 0.61 & 0.42 $\pm$ 0.35 \\ \hline
Mammo-GRG (Vicuna-13B) & 0.66 & 0.60 & 0.37 $\pm$ 0.35 \\ \hline
Mammo-GRG (Mistral-7B) & 0.57 & 0.56 & 0.34 $\pm$ 0.35 \\ \hline
\end{tabular}

\caption{\textbf{An ablation study comparing the effect of different large language models on report generation performance.} Evaluation of the Mammo-GRG framework using three underlying large language models -- Llama-3.1-8B, Vicuna-13B, and Mistral-7B—on breast imaging datasets from BU and UPMC. Performance is assessed using BLEU-1 and ROUGE-L (textual similarity metrics) and GREEN (a composite metric capturing clinical correctness and linguistic quality; mean $\pm$ s.d.). Across both datasets, Llama-3.1-8B consistently outperforms other backbones, highlighting the influence of model scale and instruction-tuning quality on radiology report generation accuracy.}
\label{tab:app_ablation_mammo_GRG_llms}
\end{table}

\subsection*{Statistical analysis for reporting numbers}
Following prior work \cite{lu2024multimodal}, we estimate 95\% confidence intervals (CIs) for all reported metrics at the patient level using non-parametric bootstrapping with 1,000 resamples. We assess statistical significance between model performances using a two-sided paired permutation test (1,000 permutations) under the null hypothesis that the two models have equal performance. In each permutation, paired prediction outcomes are randomly swapped between models to compute a new performance difference. The p-value corresponds to the proportion of permuted differences whose absolute value exceeds the observed difference.

To quantify the linear relationship between the risk scores generated by our model and the MIRAI baseline, we calculate the Pearson correlation coefficient (r). We report the p-value and the 95\% CI for this correlation. We compute the confidence interval using Fisher's z-transformation.

When reporting precision and recall, we follow previous work~\cite{yang2024limits, lipton2014optimal, seyyed2021underdiagnosis} in selecting the threshold that maximizes the F1 score. We apply this threshold optimization procedure independently for each dataset, task, and model.

\subsection*{Computing hardware and software}
We conduct all experiments in a controlled environment built with \textbf{Python 3.9.21}, \textbf{PyTorch 2.2.2}, and \textbf{CUDA 11.8}. 
We develop and fine-tune all models using \textbf{TorchVision 0.17.2}, \textbf{TorchAudio 2.2.2}, and \textbf{Transformers 4.48.0}. 
We pretrain \textbf{Mammo-FM} and \textbf{Mammo-GRG} on four \textbf{NVIDIA RTX 6000 GPUs} (48GB each). 
We apply \textbf{BitsAndBytes 0.45.2} for 8-bit quantization, \textbf{DeepSpeed 0.12.6} for distributed optimization, and \textbf{FlashAttention 2.7.3} to accelerate attention computation. 
We use \textbf{NumPy 1.26.4} and \textbf{Pandas 2.2.3} for data processing, and \textbf{Matplotlib 3.9.4} with \textbf{Seaborn 0.13.x} for visualization. 
We perform image preprocessing and augmentation with \textbf{Albumentations 2.0.0} and \textbf{OpenCV 4.10.0.84}. 
We manage model checkpoints and parameter-efficient fine-tuning through \textbf{Hugging Face Hub 0.27.1} and \textbf{PEFT 0.14.0}. 
We conduct all statistical analyses using \textbf{SciPy 1.13.1}, \textbf{StatsModels 0.14.2}, and \textbf{Scikit-learn 1.6.1}.

\subsection*{Data availability}
We use three publicly available mammography datasets in this study. The VinDr-Mammo dataset can be accessed through PhysioNet at \url{https://www.physionet.org/content/vindr-mammo/1.0.0/}. The EMBED dataset is hosted on the AWS Open Data Registry and can be accessed at \url{https://registry.opendata.aws/emory-breast-imaging-dataset-embed/}. The RSNA Breast Cancer Detection dataset is available at \url{https://www.kaggle.com/competitions/rsna-breast-cancer-detection}.  All datasets are distributed for non-commercial academic research use in accordance with their respective data use agreements and institutional review board approvals. No proprietary or private patient data is shared. 

\subsection*{Code availability}
The code used to pretrain and evaluate \textbf{Mammo-FM}, \textbf{Mammo-GRG}, and related foundation models will be made publicly available for non-commercial academic use at \url{https://github.com/batmanlab/Mammo-FM}. The repository includes detailed scripts for data preprocessing, model training, multimodal alignment, and evaluation across internal and external mammography datasets. We document all technical deep learning methods, hyperparameters, and software dependencies used in this study within the repository and supplementary materials.

\subsection*{Model availability}
We will publicly release the pretrained weights for \textbf{Mammo-FM} and \textbf{Mammo-GRG} on Hugging Face at \url{https://huggingface.co/shawn24/MammoFM}. The repository provides checkpoints for both contrastive pretraining and multimodal generative reasoning stages, enabling reproducibility and downstream adaptation. All released weights are intended for non-commercial academic research use.

\subsection*{Acknowledgements}
This work was supported in part by the Pennsylvania Department of Health; NIH Award NIH 2R01HL141813-07; NSF Career Award NSF 2443167; the Hariri Institute for Computing at Boston University; and computational resources provided by the Pittsburgh Supercomputing Center (grant TG-ASC170024). Additional support was provided by NIH/NCI U01 CA269264-01-1, “Flexible NLP toolkit for automatic curation of outcomes for breast cancer patients” (PI: I.B.).

% % % % % % % % % % % % % % % % % % % % % % % % % % % % % % 
% Generate conversation questions in Mammo-Instruct
% % % % % % % % % % % % % % % % % % % % % % % % % % % % % % 
\begin{table*}[h]
\centering
\begin{tcolorbox}[
colback=gray!5,
colframe=black,
boxrule=0.4pt,
sharp corners,
fontupper=\ttfamily\small,
width=\textwidth,
title={Prompt to generate conversation questions in Mammo-Instruct}
]
You are a radiologist and VQA dataset creator, specializing in screening breast mammogram reports. 
Using the following mammography report as background, generate 1 set of conversation-style Q\&A pairs that are visually grounded in the mammogram images. 
The set within tags <1>...</1>. Use <q>...</q> for questions and <a>...</a> for answers, ensuring most answers are derived from the images. 

Example: <1><q>Question 1</q><a>Answer 1</a><q>Question 2</q><a>Answer 2</a>...</1>.

Here are the constraints and guidelines:

1. Use the mammogram report strictly for context regarding findings and impressions. 
Do not reference the report verbatim or mention its specifics (\eg who read the exam, the software used, or the date).

2. All questions and answers must be visually driven, meaning that someone would need to look at the actual mammogram images to confirm the answer. 
Also, the question-answer must be derived from the given radiology report only, nothing else.

3. Focus strictly on the core findings or impressions. 

4. Do not include random or irrelevant questions with respect to the report. 
All questions should relate to what can be observed or concluded from the mammogram images.

5. Also include complex questions that are relevant to the report accompanied by the 2D mammogram images only. 
Provide detailed answers when answering complex questions. 
For example, give detailed examples or reasoning steps to make the content more convincing and well-organized.  

6. You can include questions about BI-RADS only if BI-RADS assessment is mentioned in the report explicitly. 
If BI-RADS is not mentioned in the report, do not include questions on the overall BI-RADS assessment.
You can include as many question and answer couples as you find appropriate. 

7. If there is any finding (mass/calcification/asymmetry etc.) mentioned, you must generate 3 questions: 
(1) what is the finding? 
(2) the corresponding laterality (left/right/bilateral), and 
(3) the corresponding view (CC or MLO) if mentioned. 
If there is no mention of views (CC/MLO), don't generate questions on views.
\end{tcolorbox}

\caption{\textbf{Prompt used for generating conversation-style question–answer pairs in Mammo-Instruct for Mammo-Instruct dataset creation.}}
\label{tab:prompt_mammo_instruct}
\end{table*}

% % % % % % % % % % % % % % % % % % % % % % % % % % % % % % 
% Generate MCQ questions in Mammo-Instruct
% % % % % % % % % % % % % % % % % % % % % % % % % % % % % % 
\begin{table*}[h]
\centering
\begin{tcolorbox}[
colback=gray!5,
colframe=black,
boxrule=0.1pt,
sharp corners,
fontupper=\ttfamily\small,
width=\textwidth,
title={Prompt for Generating Free-Response, Description, and Multiple-Choice VQA in Mammo-Instruct}
]
You are a radiologist and VQA dataset creator, specializing in screening breast mammogram reports. 
I need you to generate three sets of questions based on the given radiology report: \textbf{free response}, \textbf{description}, and \textbf{multiple-choice} questions. 
Each set should contain 3 Q/A pairs.

The free response questions should be enclosed within

\texttt{<free\_response>...</free\_response>}.  
The description questions should be enclosed within \texttt{<description>...</description>}.  
The multiple-choice questions should be enclosed within \texttt{<multiple\_choice>...</multiple\_choice>}, 
with each question having 4 choices labeled as (a), (b), (c), and (d).
Ensure that the questions are specific to the 2D breast mammogram described in the report.

\textbf{Example format:}

\texttt{<free\_response><q>What abnormalities are present in the right/left breast?</q><a>Answer</a></free\_response>}

\texttt{<description><q>Describe the findings shown in these screening mammograms.</q><a>Answer</a></description>}

\texttt{<multiple\_choice><q>What is the composition of the breast tissue observed in the mammogram? (a) Choice 1 (b) Choice 2 (c) Choice 3 (d) Choice 4</q><a>(b) Choice 2</a></multiple\_choice>}

\textbf{Guidelines:}

1. Use the mammogram report strictly for context regarding findings and impressions.  
Do not reference the report verbatim or mention its specifics (\eg who read the exam, the software used, or the date).

2. All questions and answers must be visually driven, meaning that someone would need to look at the actual mammogram images to confirm the answer.  
Also, the question–answer pairs must be derived from the given radiology report only, nothing else.

3. Focus strictly on the core findings or impressions.  

4. Do not include random or irrelevant questions. All questions must relate to observable or inferable image features.

5. Include BI-RADS–related questions only if the report explicitly mentions a BI-RADS assessment.

6. If the report mentions any finding (\eg mass, calcification, asymmetry), you must generate 3 questions:  
(1) identify the finding,  
(2) specify the laterality (left/right/bilateral), and  
(3) indicate the view (CC or MLO) if mentioned.  
If no views are mentioned, omit view-specific questions.

7. These three core questions must appear in each of the \texttt{<free\_response>}, \texttt{<description>}, and \texttt{<multiple\_choice>} sections.

\textbf{Final structure:}

% \texttt{<free\_response><q>Q1</q><a>A1</a><q>Q2</q><a>A2</a><q>Q3</q><a>A3</a></free\_response>}  
% \texttt{<description><q>Q1</q><a>A1</a><q>Q2</q><a>A2</a><q>Q3</q><a>A3</a></description>}  
% \texttt{<multiple\_choice><q>Q1 (a)...(d)</q><a>(b)</a><q>Q2 (a)...(d)</q><a>(a)</a><q>Q3 (a)...(d)</q><a>(c)</a></multiple\_choice>}

\texttt{\freeresponsestart}\\
\texttt{\quad \qstart Q1\qend\astart A1\aend}\\
\texttt{\quad \qstart Q2\qend\astart A2\aend}\\
% \texttt{\quad \qstart Q3\qend\astart A3\aend}\\
\texttt{\freeresponseend}\\[0.1ex]

\texttt{\descriptionstart}\\
\texttt{\quad \qstart Q1\qend\astart A1\aend}\\
\texttt{\quad \qstart Q2\qend\astart A2\aend}\\
% \texttt{\quad \qstart Q3\qend\astart A3\aend}\\
\texttt{\descriptionend}\\[0.1ex]

\texttt{\multiplechoicestart}\\
\texttt{\quad \qstart Q1 (a)...(d)\qend\astart(b)\aend}\\
\texttt{\quad \qstart Q2 (a)...(d)\qend\astart(a)\aend}\\
% \texttt{\quad \qstart Q3 (a)...(d)\qend\astart(c)\aend}\\
\texttt{\multiplechoicend}

\end{tcolorbox}

\caption{\textbf{Prompt used for generating structured Free-Response, Description, and Multiple-Choice question sets for Mammo-Instruct dataset creation.}}
\label{tab:prompt_multitype}
\end{table*}

\setlength{\intextsep}{4pt}
\setlength{\textfloatsep}{6pt}

\begin{table*}[t]
\centering
\begin{tcolorbox}[
    colback=gray!3,
    colframe=black,
    boxrule=0.3pt,
    sharp corners,
    width=\textwidth,
    left=3pt,
    right=3pt,
    top=4pt,
    bottom=4pt,
    fontupper=\ttfamily\footnotesize,
    before skip=0pt,
    after skip=2pt,
    title={System prompt for Mammo-GRG to generate the preliminary report after instruction tuning}
    ]

You are Mammo-GRG (Mammogram Grounded Report Generator), an AI assistant specializing in 2D screening mammograms, dedicated to providing accurate and relevant information exclusively related to 2D screening mammograms and associated medical topics.  

Every user message will already contain \textbf{four view blocks in this exact order:}  
\texttt{<LMLO>} \texttt{<LCC>} \texttt{<RMLO>} \texttt{<RCC>}  

where  
-- \texttt{<LMLO>} = Left MLO view  
-- \texttt{<LCC>} = Left CC view  
-- \texttt{<RMLO>} = Right MLO view  
-- \texttt{<RCC>} = Right CC view  

The corresponding image embedding will be inserted after each token.  
\texttt{LeftMLO} is the left mediolateral oblique view, \texttt{LeftCC} is the left craniocaudal view, \texttt{RightMLO} is the right mediolateral oblique view, and \texttt{RightCC} is the right craniocaudal view.  

\textbf{----------------------------------------------}  
\textbf{SPECIAL DIRECTIVE TOKENS}  
\textbf{----------------------------------------------}  

If the user includes one of these tokens, format your reply accordingly:  

• \texttt{<multiple\_choice>} – Reply with options (A), (B), … and then choose the best answer. 

• \texttt{<long\_answer>} – Provide a detailed paragraph-level explanation.  

• \texttt{<report\_generation>} – Output a full radiology report with \textbf{Findings} and \textbf{Impression} sections.  

If more than one directive appears, follow them in the order shown above.  

\textbf{----------------------------------------------}  
\textbf{RESPONSE GUIDELINES}  
\textbf{----------------------------------------------}  

1. Focus strictly on 2D screening-mammography findings -- tissue composition, masses, calcifications, asymmetries, distortions, overall BI-RADS category, etc.  

2. Use clear, radiology-style language; be concise and factual.  

3. If the user asks general breast-imaging knowledge (not about this exam), answer normally.  

4. If the user greets you, respond politely.  

Avoid unrelated topics and keep all responses clinically relevant.

\end{tcolorbox}
\caption{\textbf{System prompt for Mammo-GRG (preliminary report generation)} defining view-specific inputs, directive tokens, and clinical response guidelines for mammography-grounded report generation.}
\label{tab:mammo_grg_system_prompt}

\vspace{8pt} % Adds a little vertical spacing between the two tables

\begin{tcolorbox}[
    colback=gray!3,
    colframe=black,
    boxrule=0.3pt,
    sharp corners,
    width=\textwidth,
    left=3pt,
    right=3pt,
    top=4pt,
    bottom=4pt,
    fontupper=\ttfamily\footnotesize,
    before skip=0pt,
    after skip=0pt,
    title={System prompt to generate the final report from the preliminary report in the clinical grounding stage}
]

You are a radiology report assistant for screening mammography.  
Produce a \textbf{FINAL REPORT} with exactly two sections:  
1) Findings  
2) Impression  

\textbf{FOUNDATION:}  

1) Treat the PRELIMINARY REPORT as the base text. Preserve its clinically relevant content, density description, laterality, locations (\eg upper outer quadrant), and recommendations. 

2) You MUST include (carry forward) all non-contradicted statements from the preliminary report.  

3) Reconcile contradictions using the rules below.  

\textbf{RECONCILIATION RULES (VERY IMPORTANT):}  

1) If structured findings are POSITIVE for a category (mass, suspicious calcification, asymmetry), you MUST reflect that positivity in the final report:  

   -- Remove or revise any preliminary “no X” statements for that category. 
   
   -- State laterality accurately: Left, Right, or “bilaterally” ONLY if both sides are positive.
   
2) If structured findings are NEGATIVE and the preliminary report already says “no X,” keep that negative statement.  

3) Do NOT add “bilaterally” unless BOTH sides are positive.  

4) \textbf{BI-RADS POLICY (SCREENING):} You must output BI-RADS in \{0, 1, 2\} ONLY. 

   -- If any structured finding is positive/indeterminate, assign BI-RADS 0.  
   
   -- If there are no findings, assign BI-RADS 1.  
   
   -- If only benign findings are present, assign BI-RADS 2. 
   
   -- If the preliminary BI-RADS is provided, keep it if consistent; otherwise, correct it.  
   
   -- \textit{Important:} If asymmetry is positive, override any preliminary BI-RADS and assign BI-RADS 0.  

5) Do NOT mention external models or the word “classifier.”  

6) Use standard terminology (CC, MLO; laterality) and be concise and clinically appropriate. 

7) Preserve the overall style of the preliminary report while outputting ONLY the two sections.

\end{tcolorbox}
\caption{\textbf{Prompt used for the grounding stage of Mammo-GRG.} 
This instruction reconciles structured findings from Mammo-FM with the preliminary generated report to produce a clinically consistent final screening mammography report containing only \textit{Findings} and \textit{Impression} sections.}
\label{tab:mammo_grg_grounding_prompt}
\end{table*}

\begin{table*}[h]
\centering
\begin{tcolorbox}[
  colback=gray!3,
  colframe=black,
  boxrule=0.3pt,
  sharp corners,
  width=\textwidth,
  left=3pt, right=3pt, top=4pt, bottom=4pt,
  fontupper=\ttfamily\footnotesize,
  before skip=0pt, after skip=0pt,
  title={Prompt for evaluating the generated report using GREEN metric}
]
You are a breast-imaging (screening mammography) report evaluator.  
Compare the \textbf{Candidate} report to the \textbf{Reference} report and produce a \textbf{Mammography-GREEN} analysis.

Judge \textbf{only clinical content} (ignore phrasing/style). Use these \textbf{clinically significant} error categories: 

(a) \textbf{False report of a finding}: The candidate mentions a finding (\eg mass, calcification, asymmetry, architectural distortion) not in the reference.  

(b) \textbf{Missing a finding}: The candidate omits a finding present in the reference.  

(c) \textbf{Mischaracterization of a finding}: A finding appears in both reports but characteristics 
(\eg size, margins, density/descriptor, stability/interval change) differ.  

(d) \textbf{Misidentification of location/laterality}: A finding is identified but location (\eg upper outer quadrant, retroareolar, depth) or laterality (left/right/bilateral) is wrong.  

(e) \textbf{Incorrect BI-RADS score}: Final BI-RADS assessment differs. (If both omit BI-RADS, do not count an error.)

Clinically \textbf{insignificant} errors are wording differences that do \textbf{not} change meaning.

\textbf{Output requirement:} Follow the tool’s \textbf{strict JSON schema}. Be concise but clinically precise in explanations.

\textbf{Reference Report:} \{reference\_report\}

\textbf{Candidate Report:} \{candidate\_report\}
\end{tcolorbox}

\caption{\textbf{System prompt for evaluating the generated report from Mammo-GRG using GREEN metric.} The prompt instructs GPT-4o-mini to compare candidate (generated) and reference (ground truth) screening mammography reports and return clinically grounded errors under five significant categories, with strict JSON output.}
\label{tab:green_metric_prompt}
\end{table*}

\end{document}